\documentclass[journal]{IEEEtran}
\usepackage{cite}
\usepackage{amsmath}
\usepackage{algorithmic}
\usepackage{array}
\usepackage{url}
\usepackage{xcolor}
\usepackage{graphicx}
\usepackage{multirow}
\usepackage{amsmath}
\usepackage{amsthm}
\usepackage{amsfonts}
\usepackage{verbatim}
\usepackage{mathrsfs}  
\usepackage{hyperref}
\usepackage{bbding}
\usepackage[inline]{enumitem}
\DeclareMathOperator*{\argmin}{\arg\!\min}

\newtheorem{definition}{\textbf{Definition}}[section]    
\newtheorem{assumption}{\textbf{Assumption}}[section]

\usepackage[ruled,vlined]{algorithm2e}
\begin{document}
\title{Safe Model-based Off-policy Reinforcement Learning for Eco-Driving in Connected and Automated Hybrid Electric Vehicles}

\author{Zhaoxuan~Zhu,
        Nicola~Pivaro,
        Shobhit~Gupta,
        Abhishek~Gupta and Marcello~Canova.
\thanks{Z. Zhu, S. Gupta and M. Canova are with the Center for Automotive Research, The Ohio State University, Columbus, OH 43212 USA (email: zhu.1083@osu.edu; gupta.852@osu.edu; canova.1@osu.edu) }
\thanks{Nicola Pivaro is with Bending Spoons, Italy. The work is done at Center for Automotive Research, The Ohio State University. (email: nicolapivaro@gmail.com)}%
\thanks{A. Gupta is with Department of
Electrical and Computer Engineering, The Ohio State University, Columbus, OH, 43210 USA (email: gupta.706@osu.edu)}
}

\markboth{}%
{Zhu \MakeLowercase{\textit{et al.}}: Safe Model-based Deep Q-learning for Eco-Driving in Connected and Automated Hybrid Electric Vehicles}

\maketitle

\begin{abstract}
Deep Reinforcement Learning (DRL) has recently been applied to eco-driving to intelligently reduce fuel consumption and travel time.
While previous studies synthesize simulators and model-free DRL (MFDRL), this work proposes a Safe Off-policy Model-Based Reinforcement Learning (SMORL) algorithm for eco-driving. 
SMORL integrates three key components, namely a computationally efficient model-based trajectory optimizer, a value function learned off-policy and a learned safe set.
The advantages over the existing literature are three-fold. 
First, the combination of off-policy learning and the use of a physics-based model improves the sample efficiency.
Second, the training does not require any extrinsic rewarding mechanism for constraint satisfaction.
Third, the feasibility of trajectory is guaranteed by using a safe set approximated by deep generative models.

The performance of SMORL is benchmarked over 100 trips against a baseline controller representing human drivers, a non-learning-based optimal controller, a previously designed MFDRL strategy, and the wait-and-see optimal solution.
In simulation, SMORL reduces the fuel consumption by more than 21\% while keeping the average speed comparable while compared to the baseline controller and demonstrates a better fuel economy while driving faster compared to the MFDRL agent and the non-learning-based optimal controller.
\end{abstract}


\IEEEpeerreviewmaketitle

\section{Introduction}
\IEEEPARstart{W}ith the advancement in the vehicular connectivity and autonomy, Connected and Automated Vehicles (CAVs) have the potential to operate in a safer and more time- and fuel-efficient manner \cite{vahidi2018energy}. With Vehicle-to-Vehicle (V2V) and Vehicle-to-Infrastructure (V2I) communication, the controller has access to real-time look-ahead information including the terrain, infrastructure and surrounding vehicles. Intuitively, with connectivity technologies, controllers can plan a speed profile that allows the ego vehicle to intelligently pass more signalized intersections in green phases with fewer changes in speed. This problem is formulated as the eco-driving problem, which aims to minimize the weighted sum of the fuel consumption and the travel time between two designated locations by co-optimizing the speed trajectory and the powertrain control strategy \cite{sciarretta2015optimal, jin2016power}.

This field of research has experienced significant momentum in the last decade.
\cite{ozatay2014cloud,jin2016power, han2019fundamentals,sun2020optimal} address the eco-driving problem for vehicles with single power source, whereas \cite{mensing2012vehicle,guo2016optimal, olin2019reducing} study the problem with the hybrid electric powertrain architecture. 
The latter involves modeling multiple power sources and devising optimal control algorithms that can synergistically split the power demand to efficiently utilize the electric energy stored in battery. 
Maamria et al. \cite{maamria2018computation} systematically compare the computational requirements and the optimality of different formulations. 
Meanwhile, the difference also exists in the complexity of the driving scenarios. 
In \cite{ozatay2014cloud, olin2019reducing}, the eco-driving is formulated without considering the real-time traffic light variability.
\cite{jin2016power,asadi2010predictive, sun2020optimal, guo2016optimal, guo2021ecodriving} have explicitly modeled and considered Signal Phase and Timings (SPaTs) and formulate and solve the eco-driving problem with optimal control techniques. 
In this work, the eco-driving problem of Connected and Automated Hybrid Electric Vehicles (CAHEVs) with the capability of passing traffic lights autonomously is studied.

Recently, the use of Deep Reinforcement Learning (DRL) in the context of eco-driving has caught considerable attention. 
DRL provides a train-offline, execute-online methodology with which the policy is learned from the historical data or the interaction with simulated environments. 
Shi et al. \cite{shi2018application} modeled the conventional vehicles with ICE as a simplified model and implemented Q-learning to minimize the $CO_2$ emission at signalized intersections.
Li et al. \cite{li2019ecological} apply an actor-critic algorithm on the ecological ACC problem in car-following mode.
Guo and Wang \cite{guo2021integrated} proposed MPC-initialized Proximal Policy Optimization with Model-based Acceleration (PPOMA) for the problem of active signal priority control for trams.
Pozzi et al. \cite{pozzi2020ecological} designed a velocity planner considering the signalized intersection and hybrid powertrain configuration with Deep Deterministic Policy Gradient (DDPG).
Zhu et al. \cite{zhu2021deep} formulates the eco-driving problem as a Partially Observable Markov Decision Process (POMDP) and approaches it with PPO. While the strategies with Model-Free Reinforcement Learning (MFRL) in these studies show improvements in the average fuel economy and reductions in onboard computation, the methodology has a fundamental drawback.
To teach the agent to drive under complex driving scenarios while satisfying all the constraints from powertrain and traffic rules, a complex and often cumbersome rewarding/penalizing mechanism needs to be designed. 
Furthermore, under such setup, the agent learns to satisfy constraints by minimizing the expected cost. 
For scenarios that are rare yet catastrophic, the scale of the cost penalizing constraint violation needs to be significantly larger than the learning objective itself\cite{shi2018application}. 
As a result, such extrinsic rewarding mechanism increases the design period and deteriorates the final performance.

This paper proposes a safe-critical model-based off-policy reinforcement learning algorithm to solve the eco-driving problem in a connected and automated mild-HEV.
The first contribution of this work, from the eco-driving application perspective, is the elimination of the extrinsic reward design in the eco-driving problem by the design of a Model-based Reinforcement Learning (MBRL) algorithm. 
The algorithm integrates RL with trajectory optimization, which incorporates the constraints from the powertrain dynamics, vehicle dynamics, and traffic rules in a constrained optimization formulation. 
The performance of the agent is meanwhile improved by using the learned terminal cost function from the RL mechanism.

The second contribution of this work, from the RL algorithm perspective, is the development of Safe Model-based Off-policy Reinforcement Learning (SMORL), a safe-critical model-based off-policy Q-learning algorithm for systems with known dynamics. 
The algorithm has three unique features compared to the prior and current MBRL implementations. 
First, SMORL is off-policy as opposed to \cite{lowrey2018plan, karnchanachari2020practical,thananjeyan2020safety,thananjeyan2020abc}. 
While the use of the model in MBRL increases the sample efficiency \cite{wang2019benchmarking}, the collection of each individual transition becomes more computationally expensive as it commonly requires solving an online optimization problem, as opposed to a feedforward policy in MFRL.
With the use of experience replay \cite{lin1992self} in the off-policy learning, the historical data can be used to greatly reduce the overall training time.
To obtain the value function from Q function, an actor is explicitly trained as in Twin Delayed Deep Deterministic policy gradient algorithm (TD3)\cite{fujimoto2018addressing}. 
Second, the distributional mismatch between the actor and critic \cite{levine2020offline} in MBRL is explicitly addressed to improve training performance and stability with Batch Constrained Q-learning (BCQ) \cite{fujimoto2019off}.
Third, the long-term feasibility of the policy is considered by extending the safe set \cite{rosolia2019sample,thananjeyan2020safety} to a higher dimensional setting using deep unsupervised learning.

The remainder of the paper is organized as follows. 
Sec. \ref{sec: environment and formulation} presents the simulation environment and the eco-driving problem formulation. 
Sec. \ref{sec: background} introduces the preliminaries of the mathematical concepts, and Sec. \ref{sec: proposed method} presents the main algorithm SMORL. 
Sec. \ref{sec: implementation details} explains the detailed implementation of SMORL on the eco-driving problem, and Sec. \ref{sec: results} shows the training details and benchmarks the performance. 

\section{Eco-driving for CAHEVs} \label{sec: environment and formulation}
\subsection{Environment}
\begin{figure}[]
    \centering
    \includegraphics[width=\columnwidth]{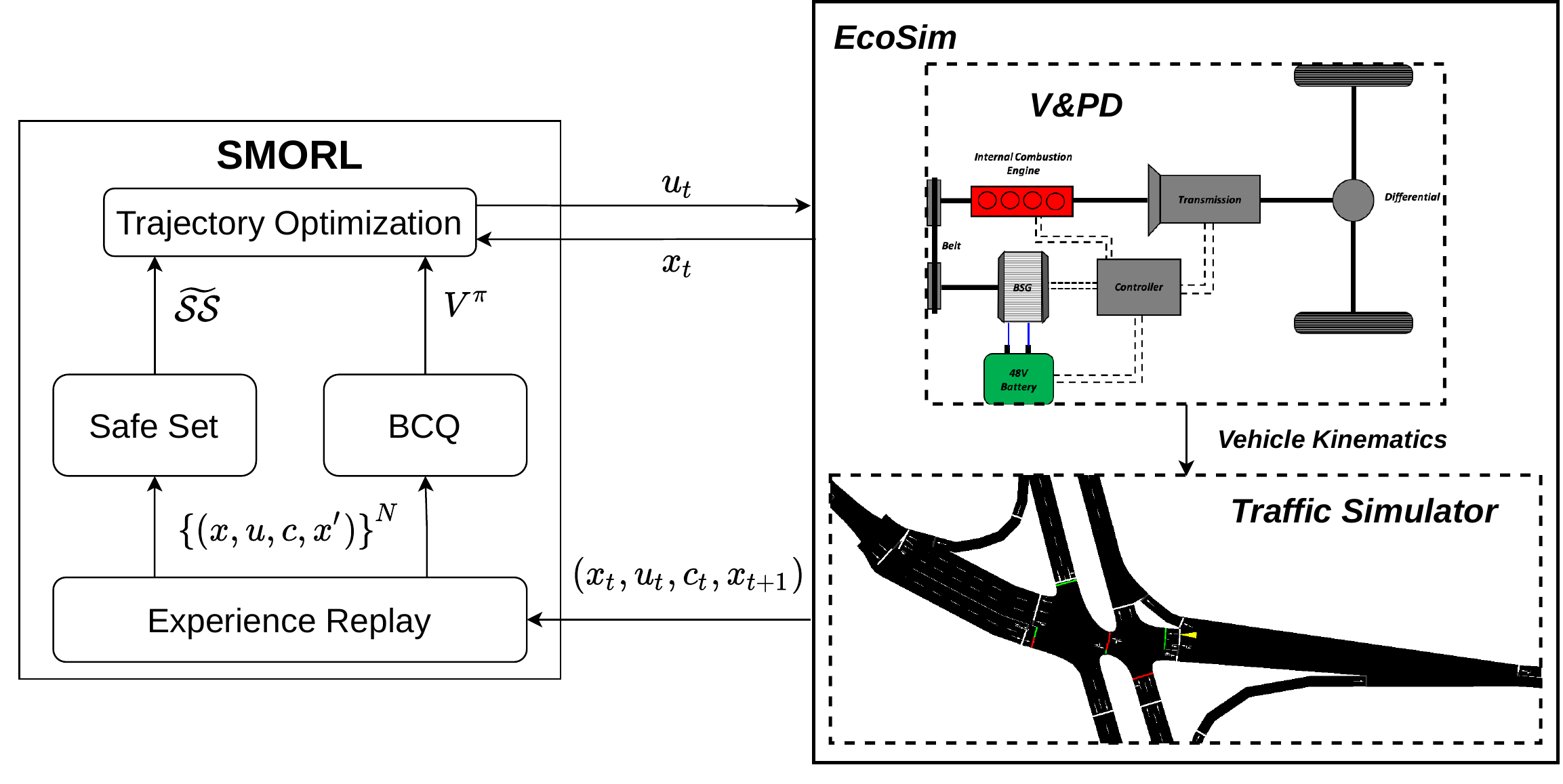}
    \caption{The Structure of The Environment Model}
    \label{fig: environment}
\end{figure}

As collecting data in real-world driving data is expensive and potentially unsafe, a model of the environment is developed for training and validation purposes. 
The environment model, named EcoSim, consists of a Vehicle Dynamics and Powertrain (VD\&PT) model and a microscopic traffic simulator. 
Fig.\ref{fig: environment} shows EcoSim and its interaction with the controller and the learning algorithm. 
The controller commands three control inputs, namely, the Internal Combustion Engine (ICE) torque, the electric motor torque and the mechanical brake torque. 
The component-level torques collectively determine the HEV powertrain dynamics, the longitudinal dynamics of the ego vehicle and its progress along the trip. 
As in \cite{asadi2010predictive}, it is assumed that the ego vehicle is equipped with Dedicated Short Range Communication (DSRC) sensors, and SPaTs from signalized intersections become available once they are within the 500 $m$ range.
The DRL agent utilizes the SPaT from the upcoming traffic light while ignoring the SPaT from any other traffic light regardless of the availability.  
Specifically, the controller receives the distance to the upcoming traffic light, its current status and its SPaT program as part of the observation. 
Finally, a navigation application with Global Positioning System (GPS) is assumed to be available on the vehicle such that the locations of the origin and destination, the remaining distance and the speed limits along the entire trip are available at every point during the trip.

\subsubsection{Vehicle and Powertrain Model} \label{sec: veh_dynamisc}
A forward-looking dynamic powertrain model is developed for fuel economy evaluation and control strategy verification. In this work, a P0 mild-hybrid electric vehicle (mHEV) is considered, equipped with a 48V Belted Starter Generator (BSG) performing torque assist, regenerative braking and start-stop functions. 

The engine is modeled as low-frequency quasi-static nonlinear maps based on steady-state engine test bench data provided by the supplier. 
The map of instantaneous fuel consumption $\dot{m}_{\mathrm{fuel}}$ is a function of engine angular velocity $\omega_{\mathrm{eng}}$ and engine torque $T_{\mathrm{eng}}$, and the maps of torque limits $T_{\mathrm{eng}}^{\mathrm{min}}$ and $T_{\mathrm{eng}}^{\mathrm{max}}$ are functions of engine angular velocity $\omega_{\mathrm{eng}}$.

The battery $SoC$ and voltage $V_{\mathrm{batt}}$ are governed by a zeroth-order equivalent circuit model shown as follows:
\begin{gather}
    I_t = \frac{V_{\mathrm{OC}}(SoC_t) - \sqrt{V_{\mathrm{OC}}^2(SoC_t) -4 R_0(SoC_t) P_{\mathrm{bsg},t}}}{2R_0(SoC_t)} \label{eq: I_batt},\\
    SoC_{t+1} = SoC_t -\frac{\Delta t}{C_{\mathrm{nom}}} (I_t + I_{\mathrm{aux}}),
\end{gather}
where $t$ is the discretized time index, and $\Delta t$ is the time discretization that is set to be $1 s$ in the study.
The power consumed by auxiliaries is modeled by a calibrated constant current bias $I_{\mathrm{aux}}$. The cell open circuit voltage $V_{\mathrm{OC}}$ and internal resistance $R_0$ are maps of $SoC$ from a battery pack supplier.  

The vehicle dynamics model is based on the road-load equation:
\begin{gather}
\begin{aligned}
	{v}_{\mathrm{veh},t+1}= & v_{\mathrm{veh}, t} + \Delta t \biggl( \frac{T_{\mathrm{out},t}-T_{\mathrm{brk},t}}{MR_\mathrm{w}} - \dfrac{C_\mathrm{d}\rho_\mathrm{a} \Omega_\mathrm{f}v_{\mathrm{veh},t}^2}{2M}  \\ 
    & - g \cos{\alpha} C_\mathrm{r}v_{\mathrm{veh},t} - g\sin{\alpha} \biggr) \label{eq: V_veh_state}
\end{aligned}
\end{gather}
Here, the four terms inside the bracket of the left-hand side are associated with the forward propulsion force, the tire rolling resistance, the aerodynamic drag, and the road grade, respectively. $T_{\mathrm{brk}}$ is the brake torque applied on wheel, $C_{\mathrm{d}}$ is the aerodynamic drag coefficient, $\rho_\mathrm{a}$ is the air density, $A_{\mathrm{f}}$ is the effective aerodynamic frontal area, $C_\mathrm{r}$ is rolling resistance coefficient, and $\alpha$ is the road grade.

Besides the aforementioned models, which are directly associated with either the states or the objective in the eco-driving Optimal Control Problem (OCP) formulation, BSG, torque converter and transmission are also modeled in the study. 
The BSG is modeled as a quasi-static efficiency map to compute the BSG torque $T_\mathrm{bsg}$ and power output $P_{\mathrm{bsg}}$. 
A torque converter model is developed to compute the losses during the traction and regeneration modes. 
The transmission model is based on a static gearbox, and its efficiency $\eta_{\mathrm{trans}}$ is scheduled as a nonlinear map of the gear number $n_\mathrm{g}$, the transmission input shaft torque $T_{\mathrm{trans}}$ and the transmission input speed $\omega_{\mathrm{trans}}$. 
The detailed mathematical models of these components can be found in \cite{zhu2021deep}.

\begin{figure}[t!]
	\centering
	\includegraphics[width=\columnwidth]{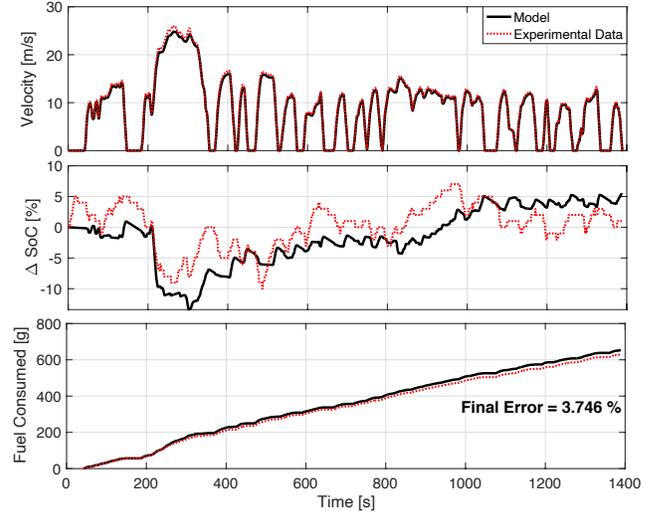}
	\caption{Validation of Vehicle Velocity, $SoC$ and Fuel Consumed over FTP Cycle.}
	\label{fig: veh_vel_soc_fuel_validation_FTP}
\end{figure}

The forward vehicle model was calibrated and validated using experimental data from a chassis dynamometer. 
Vehicle velocity, battery $SoC$, gear number, engine speed and fuel consumption were used to evaluate the model with the experimental data. 
Fig. \ref{fig: veh_vel_soc_fuel_validation_FTP} shows the sample results from model verification over the FTP-75 regulatory drive cycle. 
Results indicate that the vehicle velocity and $SoC$ are accurately predicted by the model. 
The mismatches in the battery $SoC$ can be attributed to the assumptions made in the simplified battery model such as modeling electrical auxiliary loads as a constant current bias. 
Further, the final value of the fuel consumption estimated by the model over the FTP-75 drive cycle is within 4\% of the actual fuel consumption which verifies that the model can be used for energy and fuel prediction over real-world routes.

\subsubsection{Traffic Model}
\begin{figure}[!t]
    \centering
    \includegraphics[width=1\columnwidth]{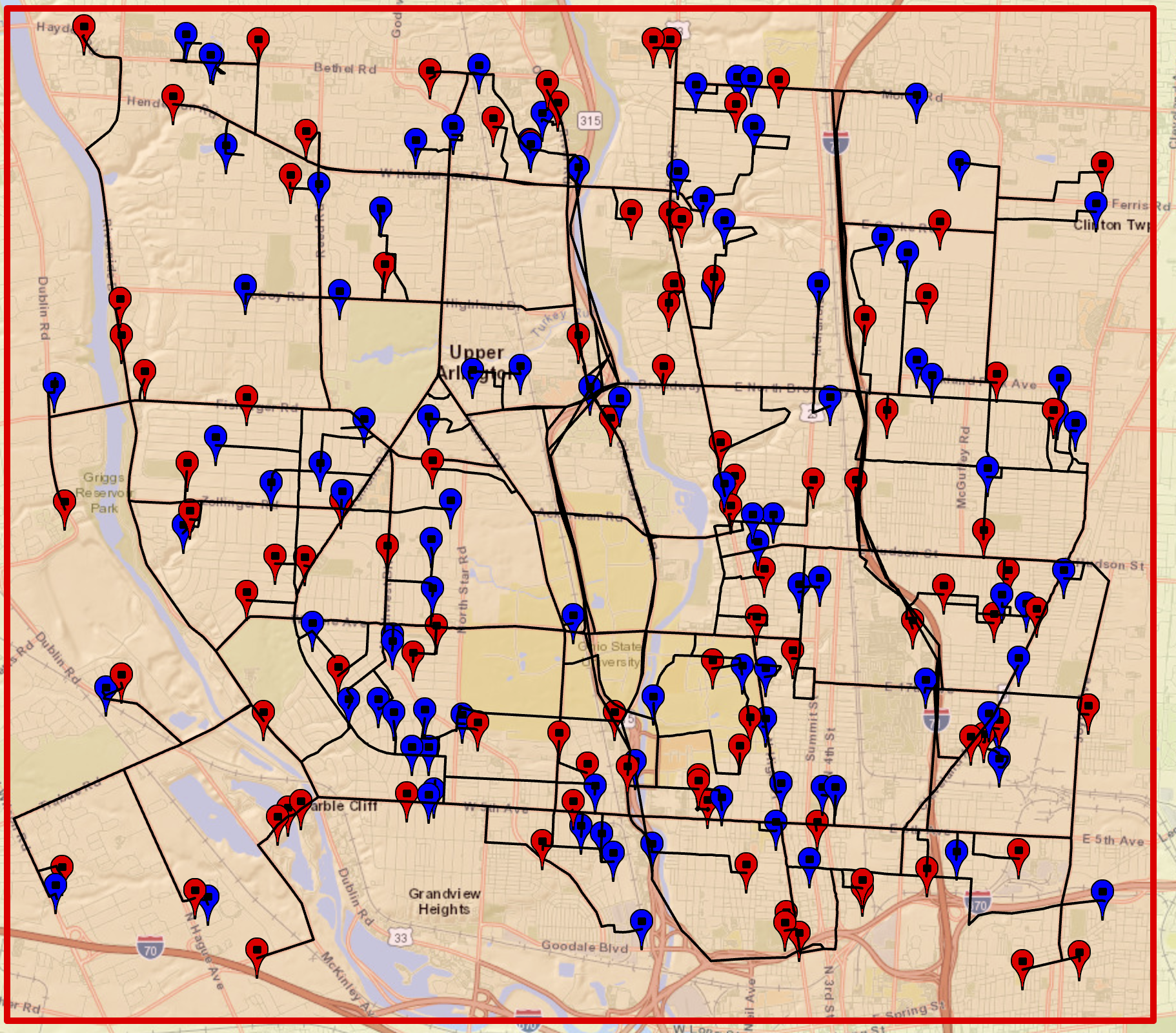}
    \caption{Map of Columbus, OH as the Traffic Environment for Training} 
    \label{fig: map_of_columbus}
\end{figure}

A large-scale microscopic traffic simulator is developed in an open source software Simulation of Urban Mobility (SUMO) \cite{SUMO2018} as part of the environment. 
To recreate realistic mixed urban and highway trips for training, the map of the city of Columbus, OH, US is downloaded from the online database OpenStreetMap \cite{OpenStreetMap}. 
The map contains the length, shape, type and speed limit of the road segments and the detailed program of each traffic light at signalized intersections. 
Fig. \ref{fig: map_of_columbus} highlights the area that is covered in the study. 
In the shaded area, 10,000 random passenger car trips, each of which the total distance is randomly distributed from 5 $km$ to 10 $km$, are generated as the training set. 
Another 100 trips, indicated by the origins (red markers) and destinations (blue markers) in Fig. \ref{fig: map_of_columbus}, are generated following the same distribution as the testing set. 
In addition, the departure time of each trip follows a geometric distribution with the success rate $p=0.01$. 
The variation among the trips used for training leads to a learned policy that is less subject to local minima and agnostic to specific driving conditions (better generalizability) \cite{heess2017emergence}.


\subsection{Optimization Formulation} \label{sec: eco-driving formulation}
In the eco-driving problem, the objective is to minimize the weighted sum of fuel consumption and travel time between two designated locations. The optimal control problem is mathematically formulated as follows:
\begin{subequations}\label{eq:OCP_formulation}
\begin{align}
    \min_{\left\lbrace u_t \right\rbrace_{t=1}^\infty}\: &\mathbb{E} \left[ \sum_{t=1}^\infty \left[\lambda\dot{m}_{\text{fuel},t} + (1-\lambda) \right]   \Delta t \cdot \mathbb{I}\left[s_t < s_{\text{total}}\right] \right]\\
     \text{where} \:&u_t = \left[T_{\text{eng},t}, T_{\text{bsg},t}, T_{\text{brk},t}\right]^T \\
     \text{s.t.} \: & SoC_{t+1} = f_{\text{batt}}(v_{\text{veh},t}, SoC_t, u_t) \label{eq:battery dynamics} \\
     & v_{\mathrm{veh},t+1} = f_{\text{veh}}(v_{\text{veh}, t}, SoC_t, u_t) \\
     & T_{\mathrm{eng}}^{\min}(\omega_{\mathrm{eng},t}) \leq T_{\mathrm{eng},t} \leq T_{\mathrm{eng}}^{\max}(\omega_{\mathrm{eng},t}) \label{eq:S2_con_Teng} \\
     & T_{\mathrm{bsg}}^{\min}(\omega_{\mathrm{bsg},t}) \leq T_{\mathrm{bsg},t} \leq T_{\mathrm{bsg}}^{\max}(\omega_{\mathrm{bsg},t}) \label{eq:S2_con_Tbsg} \\
     & I^{\min} \leq I_t \leq I^{\max} \label{eq:S2_con_I}\\
     & SoC^{\text{min}} \leq SoC_t \leq SoC^{\text{max}} \label{eq: soc_constraint}\\
     & SoC_T \geq SoC^{\mathrm{T}} \label{eq: terminal_soc_constraint}\\
     & 0 \leq v_{\mathrm{veh},t} \leq v_{\mathrm{lim},t} \label{eq: speed_limit}\\
     & (t, s_t) \notin \mathcal{S}_{\mathrm{red}}. \label{eq: traffic_constraint}
\end{align} 
\end{subequations}
Here, $\dot{m}_{\mathrm{fuel},t}$ is the instantaneous fuel consumption. 
$\lambda$ is a normalized weight on the fuel consumption. 
$\omega_{\mathrm{eng}}$ and $\omega_{\mathrm{bsg}}$ are the engine and BSG angular velocities, respectively, and they are static functions of vehicle speed $v_{\mathrm{veh}}$ and gear number $n_\mathrm{g}$.
$f_{\mathrm{batt}}$ and $f_{\mathrm{veh}}$ are the battery and vehicle dynamics, respectively, introduced in Sec. \ref{sec: veh_dynamisc}. 
Eqn. \eqref{eq:S2_con_Teng} to \eqref{eq:S2_con_I} are the constraints imposed by the powertrain components. 
Eqn. \eqref{eq: soc_constraint} and Eqn. \eqref{eq: terminal_soc_constraint} are the constraints on the instantaneous battery $SoC$ and terminal $SoC$ for charge sustaining, respectively. 
Here, the subscript $T$ represents the time at which the vehicle reaches the destination.
$SoC^\mathrm{min}$, $SoC^\mathrm{max}$ and $SoC^\mathrm{T}$ are commonly set to 30\%, 80\% and 50\% \cite{olin2019reducing,deshpande2021real}. 
Eqn. \eqref{eq: speed_limit} and \eqref{eq: traffic_constraint} are the constraints imposed by traffic conditions. 
The set $\mathcal{S}_{\mathrm{red}}$ represents the set in which the traffic light at the certain location is in red phase \cite{zhu2021gpu}. 
The problem is formulated as an infinite horizon problem in which the stage cost becomes zero once the system reaches the goal set, i.e. the traveled distance $s_t$ is greater than or equal to the total distance of the trip $s_{\mathrm{total}}$ while keeping the terminal $SoC_T$ greater than or equal to $SoC^{\mathrm{T}}$. 
In addition, anytime that the vehicle violates the traffic light constraints, i.e. Eqn. \eqref{eq: traffic_constraint}, the trip is considered a failure and the goal set is not reached. 

To solve the aforementioned optimization formulation as an OCP, a MBRL algorithm is proposed, and the preliminaries of the algorithm are included in the next section.  

\section{Preliminaries on MBRL}\label{sec: background}
The nonlinear, stochastic, time-invariant system is considered in this work:
\begin{equation}
\begin{aligned}
    x_{t+1} &= f(x_t,u_t,w_t)\\
    x_t &\in \mathcal{X} \subseteq \mathbb{R}^n, \: t \in \mathbb{N}_{+}\\
    u_t &\in \mathcal{U}(x_t) \subseteq \mathbb{R}^m, \: t \in \mathbb{N}_{+}\\
    w_t &\in \mathcal{W} \subseteq \mathbb{R}^p,  \: t \in \mathbb{N}_{+}. \label{eq: system dynamics}
\end{aligned} 
\end{equation}
Here, $x_t$, $u_t$ and $w_t$ are the state, control, and uncertainty at time $t$. 
$\mathcal{X}$ and $\mathcal{U}$ are the feasible sets for states and inputs, respectively.
The uncertainties are assumed to be independent and identically distributed (i.i.d.).

Let $\pi:\mathcal{X} \rightarrow \mathcal{U}$ be a feasible deterministic policy and $\Pi$ be the set of all feasible deterministic policies. 
The objective of the OCP is to reach the goal set $\mathcal{G}\subseteq \mathbb{R}^n$ while finding the optimal policy $\pi^*$ that minimizes the expectation of the discounted sum of the costs defined as follows:
\begin{equation}
\begin{gathered}
    \pi^* = \argmin_{\pi \in \Pi}\eta(\pi), \; \text{where} \\
    \eta(\pi) = \mathbb{E}_{w_t}\left[ \sum_{t=0}^\infty  \gamma^{t}c\left(x_t,u_t\right)\right],  \\
    \text{where } u_t = \pi(x_t).
\end{gathered}
\end{equation}
Here, $\gamma$ is the discount factor that prioritizes the immediate rewards and ensures the sum over the infinite horizon remains finite. 

As in \cite{thananjeyan2020abc}, the following assumption is made. 
\begin{assumption}[Costs]
The cost is zero for the states inside the goal set $\mathcal{G}$ and positive for the states outside, i.e. $\exists \epsilon > 0$ such that $c(x,u)>\epsilon\mathbb{I}_{\mathcal{G}^C(x)}$ where $\mathbb{I}$ is the indicator function and $\mathcal{G}^C$ is the complement of the goal set $\mathcal{G}$.
\end{assumption}

As in\cite{borrelli2017predictive,rosolia2019sample,thananjeyan2020abc}, the following definitions are given.
\begin{definition}[Robust Control Invariant Set]
A set $\mathcal{C}\subseteq\mathcal{X}$ is said to be a robust control invariant set for the system Eqn. \eqref{eq: system dynamics} if for all $x(t) \in \mathcal{C}$, there exists a $u(t)\in \mathcal{U}$ such that $f(x(t),u(t), w(t))\in \mathcal{C}$, for all $w(t)\in \mathcal{W}$ and $t\in \mathbb{N}_+$.
\end{definition}

\begin{definition}[Robust Successor Set $\textit{Suc}(\mathcal{S})$]
For a given set $\mathcal{S}$, its robust successor set $\text{Suc}(\mathcal{S})$ is defined as 
\begin{align}
\begin{aligned}
    \text{Suc}(\mathcal{S})=&\left\lbrace x'\in \mathbb{R}^n: \exists x \in \mathcal{S}, \exists w \in \mathcal{W} \right. \\
    & \quad \quad \quad \quad \: \left. \text{such that } x' = f(x, \pi(x), w) \right\rbrace.
\end{aligned}
\end{align}
\end{definition}

\begin{definition}[Robust Reachable Set $\mathcal{R}_N(x_0^j)$]For a given initial state $x_0^j$, the N-step robust reachable set $\mathcal{R}_N(x_0^j)$ of the system defined in Eqn. \eqref{eq: system dynamics} in a closed loop policy $\pi$ at iteration $j$ is defined recursively as
\begin{equation}
\begin{aligned}
    \mathcal{R}_{i+1}^{\pi}(x_0^j) &= \text{Suc}(\mathcal{R}_i^{\pi}(x_0^j)) \cap \mathcal{X},\\
    \mathcal{R}_0^\pi(x_0^j) &= x_0^j,
\end{aligned}
\end{equation}
where $i = 0, 1, \dots, N-1$.
\end{definition}

\begin{definition}[Safe Set]
The safe set $\mathcal{SS}^j$ contains the full evolution of the system at iteration j, 
\begin{align}
    \mathcal{SS}^j=\left\lbrace \bigcup_{k=0}^\infty \mathcal{R}_k^{\pi} (x_0^j) \bigcup \mathcal{G}\right\rbrace. \label{eq: safe set}
\end{align}
\end{definition}
As shown in \cite{rosolia2019sample}, the exact form of the safe set in \ref{eq: safe set} is a robust control invariant set. As calculating its exact form is intractable, especially for high dimensional nonlinear system, it is, in practice, approximated as 
\begin{align}
    \widetilde{\mathcal{SS}}^j = \bigcup_{k\in \mathcal{M}^j} x^k,
\end{align}
where $x^k=\lbrace x_t^k:t\in \mathbb{N}_+\rbrace$ is the trajectory at iteration $k$, and $\mathcal{M}^j = \lbrace k\in [0,j): \lim_{t\rightarrow \infty} x_t^k\in\mathcal{G}\rbrace$ is the set of indices of which the trajectories were successfully driven to the goal. As the safe set in this work is constantly evolving during training, the iteration index $j$ will be neglected in the remaining work. 

For any policy $\pi$, the value function $V^\pi:\mathcal{X}\rightarrow \mathbb{R}$, the Q function $Q^\pi:\mathcal{X}\times \mathcal{U} \rightarrow \mathbb{R}$ and the advantage function $A^\pi:\mathcal{X}\times \mathcal{U} \rightarrow \mathbb{R}$ are defined as follows:
\begin{align}
    V^{\pi}(x_t) &= \begin{cases}
        \mathbb{E}_{\pi}\left[\displaystyle{\sum_{i=t}^\infty}\gamma^{i-t}c\left(x_i,u_i\right)|x_t\right], & x_t \in \mathcal{SS}\\
        \infty, & \text{otherwise.}
    \end{cases}\\
    Q^{\pi}(x_t,u_t) &= \begin{cases} \mathbb{E}_{\pi}\left[\displaystyle{\sum_{i=t}^\infty}\gamma^{i-t}c\left(x_i,u_i\right)|x_t,u_t\right], & \begin{aligned}&x_t \in \mathcal{SS}\\ &u_t \in \mathcal{U} \end{aligned} \\
        \infty & \text{otherwise.} 
    \end{cases}\\
    A^{\pi}(x_t,u_t) &=Q^{\pi}(x_t,u_t) - V^{\pi}(x_t).
\end{align}

\section{Proposed Method} \label{sec: proposed method}
In this work, an off-policy model-based deep reinforcement learning algorithm with an approximated safe set is proposed. At any given time $t$ during policy execution, the following trajectory optimization problem with a receding horizon of $H$ steps is solved:
\begin{equation}
\begin{aligned}
    \min_{\left\lbrace \tilde{u}_k \right\rbrace_{k=t}^{t+H-1}} & \mathbb{E}\left[ \sum_{k=t}^{t+H-1} \gamma^{k-t}c(\tilde{x}_k,\tilde{u}_k) + \gamma^H V^{\pi} (\tilde{x}_{k+H}) \right] \\
    \text{s.t.} \:& \tilde{x}_{k+1} = f(\tilde{x}_k,\tilde{u}_k,w_k)\\
    & \tilde{x}_t = x_t \\
    & \tilde{x}_k \in \mathcal{X}, \: k = t, \dots, t+H-1 \\
    & \tilde{x}_{t+H} \in \widetilde{\mathcal{SS}}\\
    & \tilde{u}_k \in \mathcal{U}, \: k = t, \dots, t+H-1,
\end{aligned} \label{eq: traj optim val func}
\end{equation}
where $\tilde{x}$ and $\tilde{u}$ are the variables for states and control actions in the predicted trajectory. 
Compared to the formulation in \cite{lowrey2018plan}, the state $\tilde{x}_k$ and action $\tilde{u}_k$ are explicitly constrained to be within the feasible region in the receding horizon and the terminal state $\tilde{x}_{t+H}$ to be within the safe set $\widetilde{\mathcal{SS}}$. With the presence of uncertainties in the dynamic system, solving the exact form of the above stochastic optimization problem can be challenging. In \cite{chua2018deep, thananjeyan2020safety, thananjeyan2020abc}, Cross Entropy Method (CEM) \cite{chua2018deep} is used to solve the problem with unknown dynamics as a chance constraint problem. In Section \ref{sec: eco-driving formulation}, techniques will be discussed to simplify and solve the optimization in the eco-driving problem.

As most of the model-based deep reinforcement learning methods with trajectory optimization in literature learn the value function as the terminal cost for the MPC \cite{chua2018deep, lowrey2018plan, thananjeyan2020abc, karnchanachari2020practical}, the learning algorithm becomes on-policy. 
While the trajectory optimization increases the sample efficiency and helps exploration\cite{lowrey2018plan}, solving the trajectory optimization problem makes each data sample more computationally expensive. 
As a result, the training wall time is not necessarily reduced. 
In this work, the off-policy Q-learning \cite{watkins1989learning} is instead proposed. 
To use the learned Q function in trajectory optimization, the following equation needs to be solved:
\begin{align}
    \min_{\left\lbrace \tilde{u}_k \right\rbrace_{k=t}^{t+H}} & \mathbb{E}\left[ \sum_{k=t}^{t+H-1} \gamma^{k-t}c(\tilde{x}_k,\tilde{u}_k) + \gamma^H Q^{\pi}_\theta (\tilde{x}_{t+H},\tilde{u}_{t+H}) \right],\label{eq: traj optim q func}
\end{align}
where $Q_\theta^{\pi}$ is the approximated $Q$ function parametrized by $\theta$.
Compared to solving Eqn.\eqref{eq: traj optim val func}, solving Eqn. \eqref{eq: traj optim q func} requires one extra computational step:
\begin{align}
    V^\pi(\tilde{x}_{t+H}) = \min_{\tilde{u}_{t+H}} Q_\theta^{\pi} (\tilde{x}_{t+H},\tilde{u}_{t+H}). \label{eq: q to v}
\end{align}
Depending on the dimension of the problem, solving Eqn. \eqref{eq: q to v} can be computationally intractable, especially for online control. 
Several algorithms, e.g. DDPG \cite{lillicrap2015continuous}, TD3 \cite{fujimoto2018addressing} and dueling network \cite{wang2016dueling} are proposed to obtain the value function from the Q function.
In this work, the off-policy actor-critic algorithm TD3 is used since it reduces the overestimation and is shown to be more stable than DDPG. 
Specifically, with the sample $(x_j, u_j, c_j, x'_j)$ from the experience replay buffer $\mathcal{D}$ \cite{lin1992self}, the target for the Q function during training is constructed as follows, 
\begin{gather}
    y_j = c_j + \gamma \max_{i=1,2}Q_{\theta'_i}(x'_j,u'_j),\\
    u'_j = \pi_{\phi'}(x'_j) \label{eq: td3_actor}.
\end{gather}
Here, $Q_{\theta_1}$ and $Q_{\theta_2}$ are two independently trained critic networks.
$Q_{\theta'_1}$ and $Q_{\theta'_2}$ are the corresponding target networks. 
$\pi_{\phi}$ and $\pi_{\phi'}$ are the actor network and its target network, respectively. 
The critics are then updated following
\begin{equation}
    \begin{gathered}
    \theta_i \leftarrow \theta_i - \alpha \nabla_{\theta_i} \left[\dfrac{1}{N}\sum_{j=1}^N\left(y_j-Q_{\theta_i}(x_j,u_j)\right)^2 \right], \: i = 1, 2, \label{eq: critic update} \\
    \left\lbrace(x_j, u_j, c_j, x'_j) \sim \mathcal{D} \right\rbrace_{j=1}^N.
    \end{gathered}
\end{equation}
where $\alpha$ is the learning rate, and $N$ is the batch size. 

In the off-policy learning algorithm used here, the behavior policy is the trajectory optimization where state and action constraints within the receding horizon are satisfied thanks to the constrained optimization formulation. 
However, the trained actor $\pi_\phi$ makes decisions solely based on the Q function. 
The resulting mismatch between the distribution of state-action pairs induced by the actor $\pi_\phi$ and that collected by the behavior policy results in extrapolation error leading to unstable training \cite{levine2020offline}. 
In the eco-driving problem, the trajectory optimization ensures the power is solely generated from ICE when the battery $SoC$ is at the lower limit $SoC^{\mathrm{min}}$. 
Accordingly, no state-action pair resembling low $SoC$ and high motor torque can be collected, which leads to extrapolation in the Q function near the region. 
The error can eventually cause unstable training or inferior performance.

To address the extrapolation error induced by the mismatch in distributions, Batch Constrained Q-learning (BCQ) \cite{fujimoto2019off} originally proposed for offline reinforcement learning is used. 
Here, a generative model, specifically a Variational Autoencoder (VAE) \cite{kingma2013auto}, $G_\omega(x)$ is trained to resemble the state-action distribution in the experience replay buffer. 
The background on VAE and the training objective are covered in Appendix \ref{appendix: vae}.
Note that samples from the generative model $a' \sim G_\omega(x')$ should ideally match the distribution collected by the behavior policy. 
Instead of selecting action following Eqn. \eqref{eq: td3_actor}, the action is now selected as
\begin{equation}
\begin{gathered}
    u'_j = \argmin_{u_{j,k} + \xi_\phi(x_j, u_{j,k}, \Phi)} \left[\max_{i=1,2} Q_{\theta_i'}(x'_j,u_{j,k} + \xi_\phi(x_j, u_{j,k}, \Phi))\right],\\
    \left\lbrace u_{j,k} \sim G_\omega(x'_j) \right\rbrace_{k=1}^n.
\end{gathered} 
\end{equation}
Here $n$ is the hyperparameter that is the number of actions sampled from the generative model. 
The action $u'$ used for the target value for the Q function is selected as the best among the $n$ sampled ones. 
Note that there is no longer an actor network mapping from state to action. 
Instead, to ensure the agent can learn on top of the actions sampled from the generative model imitating the behavior policy from the experience buffer, a perturbation network $\xi_\phi$ whose output is clipped between $[-\Phi,\Phi]$ is trained. 
The perturbation network $\xi_{\phi}$ is updated by deterministic policy gradient theorem from \cite{silver2014deterministic} as 
\begin{align}
    \phi \leftarrow \phi - \alpha \nabla_\phi\left[\dfrac{1}{N} \sum_{j=1}^N Q_{\theta_1}(x_j, u_j + \xi_\phi(x_j, u_j, \Phi))\right]. \label{eq: perturb update} 
\end{align} 
To reduce the accumulating error from bootstrapping, all the target networks are updated with a slower rates as
\begin{align}
    \theta'_i &\leftarrow \tau \theta_i + (1-\tau) \theta'_i, \quad i = 1, 2, \label{eq: critic target update}\\
    \phi' &\leftarrow \tau \phi + (1-\tau) \phi', 
\end{align}
where $\tau$ is a constant on the order of $10^{-3}$ to $10^{-1}$. 

In Eqn. \eqref{eq: traj optim val func}, the terminal rollout state is constrained within the safe set.
Since the safe set is an approximation to the robust control invariatn set, the constrain ensures that there exists policies that can safely drive the terminal state to the goal set.
In \cite{thananjeyan2020safety, thananjeyan2020abc}, the safe set is approximated by kernel density estimation, which typically works well only for problems in low dimensions. 
Here, we extend the approximation to high-dimensional setting by using deep generative models. 
Following the notion in \cite{thananjeyan2020safety}, the safe set is approximated as 
\begin{align}
    \widetilde{\mathcal{SS}} = \lbrace x:p_\psi(x) \geq \delta \rbrace,
\end{align}
where $p_\psi:\mathcal{X}\rightarrow[0,1]$ is the probability that a state is inside the safe set parametrized by $\psi$, and the constant $\delta$ regulates how exploratory the controller is.
Note that the generative model used for the safe set approximation needs to model the probability explicitly and can be slow in sampling, whereas the generative model resembling the distribution of state-action pairs in the experience replay needs to be fast in sampling while the explicit probability is not required. 

Due to the aforementioned consideration, the autoregressive model with Long Short-term Memory (LSTM) \cite{karpathy2015visualizing} is used. The description of the model as well as the training objective is included in Appendix \ref{appendix: autoregressive lstm}.
In Sec.\ref{sec: implementation details}, the use of the autoregressive model in the application of eco-driving is motivated as the dimension of the problem can get large once the future conditions are sampled discretely.

In summary, Safe Model-based Off-policy Reinforcement Learning (SMORL) is proposed. The algorithm builds on SAVED \cite{thananjeyan2020safety} and extends it to be an off-policy algorithm with the methods proposed in BCQ. The detailed step-by-step algorithm is included in Algorithm \ref{algorithm: smorl}.

\begin{algorithm*}[]
\SetAlgoLined
 Initialize Q-networks $Q_{\theta_1}$, $Q_{\theta_2}$ independently, and duplicate target networks $Q_{\theta'_1}$, $Q_{\theta'_2}$. \\
 Initialize the perturbation network $\xi_\phi$, its target network $\xi_\phi'$ and VAE $G_\omega=\left\lbrace E_{\omega_1}, D_{\omega_2} \right\rbrace$.\\
 Initialize the experience replay buffer $\mathcal{D}$.\\
 Collect $N_0$ successfully executed trajectories with a baseline controller and initialize the safe set $\widetilde{\mathcal{SS}}$.\\
 \For{$n_{iter}\in{1,\dots,N_{iter}}$}{
    \While{$j^{th}$ trajectory NOT finished}{
    Select control action $u_t$ by solving trajectory optimization in Eqn. \eqref{eq: traj optim val func}.\\
    Sample mini-batch of $N$ transitions $(x, u, c, x')$ from $\mathcal{D}$.\\
    For each transition, sample $n$ actions $u'_j$ from $G_\omega(x')$ and $n$ perturbations from $\xi_\phi(x',u',\Phi)$. \\
    Update the critic networks $Q_{\theta_1}$, the target networks $Q_{\theta_2}$ following Eqn. \eqref{eq: critic update} and $Q_{\theta'_1}$, $Q_{\theta'_2}$ following Eqn. \eqref{eq: critic target update}. \\
    Update perturbation network $\xi_\phi$ following Eqn. \eqref{eq: perturb update}. \\
    Update VAE $G_\omega$ by maximizing Eqn. \eqref{eq: vae update}. 
    }
    \If{$x_T\in\mathcal{G}$}{
    Push the trajectory $\left\lbrace (x_t,u_t,c_t,x_{t+1}) \right\rbrace^T$ to $\mathcal{D}$.\\
    Update the safe set $\widetilde{\mathcal{SS}}$ with minibatchs sampled from $\mathcal{D}$ following Eqn. \eqref{eq: autoregressive update}. 
    }
 }
 \caption{Safe Model-based Off-policy Reinforcement Learning (SMORL)}
 \label{algorithm: smorl}
\end{algorithm*}

\section{Implementation Details} \label{sec: implementation details}
\subsection{Trajectory Optimization}
Specific to the eco-driving problem, the state vector $x_t$ is defined as a vector with 88 states. 
A description of the states are listed in Tab. \ref{tab: state_action_space}. 
Here, the first seven elements of the state vector are the battery $SoC$, the vehicle speed $v_{\mathrm{veh}}$, the current speed limit $v_{\mathrm{lim}}$, the next speed limit $v_{\mathrm{lim}}'$, the distance to the next speed limit $s_{\mathrm{lim}}$, the distance to the upcoming traffic light $s_{\mathrm{tls}}$ and the total remaining distance $s_{\mathrm{rem}}$. 
The remaining 81 elements are the sampled upcoming traffic light status in the next 80 seconds $x_{\mathrm{tfc}}$. 
For example, if the upcoming traffic light has 20 seconds remaining for the current red phase and will remain in green for the rest of the 80 seconds, the first 21 elements of the sampled upcoming traffic light status are 0, and the rest are set to 1. 
Compared to the manually extracted feature representation in \cite{zhu2021deep}, the sampled representation reduces the discontinuity and results in a better performance. 

\begin{table}[]
    \centering
    \caption{The State and Action Spaces of the Eco-driving Problem}
    \label{tab: state_action_space}
    \begin{tabular}{c|c|p{5cm}}
    & Variable & \multicolumn{1}{c}{Description}\\ [3pt]
    \hline
    \multirow{8}{*}{$\mathcal{X}$} & $SoC \in \mathbb{R}$ & Battery $SoC$ \\[3pt]
    & $v_{\text{veh}}\in \mathbb{R}$ & Vehicle velocity\\[3pt]
    & $v_{\text{lim}}\in \mathbb{R}$ & Speed limit at the current road segment\\[3pt] 
    & $v'_{\text{lim}}\in \mathbb{R}$ & Upcoming speed limit\\[3pt]
    & $d_{\text{tfc}}\in \mathbb{R}$ & Distance to the upcoming traffic light\\[3pt]
    & $d'_{\text{lim}}\in \mathbb{R}$ & Distance to the road segment of which the speed limit changes\\[3pt]
    & $d_{\text{rem}}\in \mathbb{R}$ & Remaining distance of the trip\\[3pt]
    & $x_{\text{tfc}} \in \left\lbrace 0, 1 \right\rbrace^{81}$ & Sampled status of the upcoming traffic light\\[3pt]
    \hline
    \multirow{3}{*}{$\mathcal{U}$} & $T_{\text{eng}}\in \mathbb{R}$ & Engine torque\\[3pt]
    & $T_{\text{bsg}}\in \mathbb{R}$ & Motor torque \\[3pt]
    & $T_{\text{brk}}\in \mathbb{R}$ & Equivalent brake torque
    \end{tabular}
\end{table}

As the vehicle considered in this study is assumed equipped with connected features, e.g. advanced mapping and V2I connectivity, and surrounding vehicles are not included in the study, it is assumed that the ego vehicle can deterministically predict the uncertainties from driving conditions within the receding horizon in this study. Sun et al. \cite{sun2020optimal} suggest by formulating the problem as a chance constraint or a distributionally robust optimization problem, uncertainties in SPaT can be considered without additional computational load.  

In Eqn. \eqref{eq:OCP_formulation}, the receding horizon $H$ is in the time domain. While it is easier to incorporate the time-based information such as SPaT received from V2I communication in the time domain, an iterative dynamic look-ahead process is required to process any distance-based route feature, such as speed limits, grade, traffic light and stop sign locations. 
For example, the controller requires the speed limits as the constraints to generate speed trajectory while the speed limits can change based on the distance traveled by the speed trajectory. 
In this study, the value and Q functions are learned in the time domain for ease of integration with the time-based traffic simulator, while the trajectory optimization is conducted in the spatial domain.

As SPaTs and speed limits do not depend on the decision made by the ego vehicle in the spatial domain, they are incorporated into the optimization problem as constraints, and only the vehicle speed, battery $SoC$ and the time at which the vehicle reaches the given distance are considered as the state in the trajectory optimization. 
Define the optimization state $z\in \mathcal{Z} \subseteq \mathbb{R}^3$ as
\begin{align}
    z_s=\begin{bmatrix}v_{\mathrm{veh},s}, SoC_s, t_s\end{bmatrix}^T.
\end{align}
Here, $s$ is the index in the discretized spatial domain with $\Delta s = 10 m$, and the dynamics of $z$ in the time and spatial domains are converted following 
\begin{align}
    \dfrac{\Delta z}{\Delta s} = \dfrac{\Delta z}{\Delta t}\dfrac{\Delta t}{\Delta s} = \dfrac{\Delta z}{\Delta t}\dfrac{1}{v_{\mathrm{veh}}}.
\end{align}

As a result, the trajectory optimization is formulated as
\begin{subequations}
\begin{align}
 \min_{\left\lbrace \tilde{u} \right\rbrace_{k=s_t}^{s_t+H_s-1}} \: &\sum_{k=s_t}^{s_t+H_s-1} \gamma^{t_k} c(\tilde{z}_k,\tilde{u}_k) + \gamma^{t_{H_s}} V^\pi(\mathcal{G}(x_t,z_{H_s}))\\
\text{where:} \: & \notag \\
c(\tilde{x}_k, \tilde{u}_k) = &\left( \lambda \dot{m}_{\mathrm{fuel},k} + (1-\lambda) \right)   \dfrac{\Delta s}{v_{\mathrm{veh},k}} \cdot\mathbb{I}\left[s_k < s_{\mathrm{total}}\right]\\
\text{s.t.} \: & SoC_{k+1}  = f_{\mathrm{batt},s}\left(\tilde{z}_k, \tilde{u}_k\right) \\
& v_{\mathrm{veh},k+1} = f_{\mathrm{veh},s}\left(\tilde{z}_k, \tilde{u}_k\right)\\
& T_{\mathrm{eng}}^{\min}(\omega_{\mathrm{eng},k}) \leq T_{\mathrm{eng},k} \leq T_{\mathrm{eng}}^{\max}(\omega_{\mathrm{eng},k}) \\
& T_{\mathrm{bsg}}^{\min}(\omega_{\mathrm{bsg},k}) \leq T_{\mathrm{bsg},k} \leq T_{\mathrm{bsg}}^{\max}(\omega_{\mathrm{bsg},k}) \\
& I^{\min} \leq I_k \leq I^{\max} \\
& SoC^{\text{min}} \leq SoC_k \leq SoC^{\text{max}}\\
& 0 \leq v_{\mathrm{veh},k} \leq v_{\mathrm{lim},k}\\
& (t_k, s_k) \notin \mathcal{S}_{\mathrm{red}}\\
& \mathcal{G}(x_t, z_{H_s}) \in \widetilde{\mathcal{SS}}.
\end{align}
\end{subequations}
Here, $s_t$ is the spatial index corresponding to the distance the ego vehicle has traveled at the time $t$. 
$H_s=20$ is the prediction step in the spatial domain, making the total prediction horizon 200 $m$. 
$\mathcal{G}: \mathcal{X} \times \mathcal{Z} \rightarrow \mathcal{X}$ is the function that takes the full state $x_t$ and the terminal optimization state $z_{H_s}$ and determines the predicted terminal full state $\tilde{x}_{t+t_{H_s}}$. 
For example, suppose there are 15 seconds left in the current green phase and $t_{H_s}$ in the optimization state is 10 seconds, i.e. it takes 10 seconds for the ego vehicle to travel the future 200 $m$, there will be 5 seconds left in the current green phase at the end of the prediction horizon. 
The trajectory optimization problem is solved by Deterministic Dynamic Programming (DDP) \cite{sundstrom2009generic}. 
The optimal deterministic policy $\mu_k^*: \mathcal{Z} \rightarrow \mathcal{U}$, $k=1, 2, \dots, H_{s}-1$, along with the optimal cost-to-go function $\mathcal{J}_k: \mathcal{Z} \rightarrow \mathbb{R}$, $k=1, 2, \dots, H_s$ can be calculated through backward recursion as
\begin{subequations}\label{eq: dp update}
\begin{gather}
    \mathcal{J}_{H_s}(z) = V^\pi\left(\mathcal{G}(x_t,z)\right) + \mathcal{P}_N(z) \label{eq: dp terminal cost}, \\
    \mathcal{F}_k(z,u) = c(z,u) + \mathcal{P}_k(z) + \mathcal{J}_{k+1}(f_k(x,u)),\\
    \mu_k^*=\argmin_{\mu_k} \mathcal{F}_k(z, \mu_k(z)) \label{eq: dp optimal policy},\\
    \mathcal{J}_{k}(z) = \mathcal{F}_k(z, \mu^*_k(z)) \label{eq: dp optimal value}.
\end{gather}
\end{subequations}
Here, $\mathcal{F}: \mathcal{Z}\times \mathcal{U}\rightarrow \mathbb{R}$ is the cost-to-go associated with a given immediate action and then the optimal policy.
$\mathcal{P}_k:\mathcal{Z}\rightarrow\mathbb{R}$ and $\mathcal{P}_N:\mathcal{Z}\rightarrow\mathbb{R}$ are penalty functions introduced to ensure no constraint violation in the predicted trajectory.

Solving Eqn. \eqref{eq: dp update} is computationally intensive yet highly parallelizable. 
Considering onboard GPU is readily available nowadays on self-driving vehicles, a real-time capable CUDA-based Parallel DDP  (PDDP) solver in \cite{zhu2021gpu} is used in this work.
In the cases where the stochasticity within the prediction horizon cannot be ignored, other gradient-free optimization methods, such as CEM or random shooting method \cite{nagabandi2018neural}, can be used as the trajectory optimizer. 

\subsection{Q Learning}
\begin{figure}[!t]
    \centering
    \includegraphics[width=1\columnwidth]{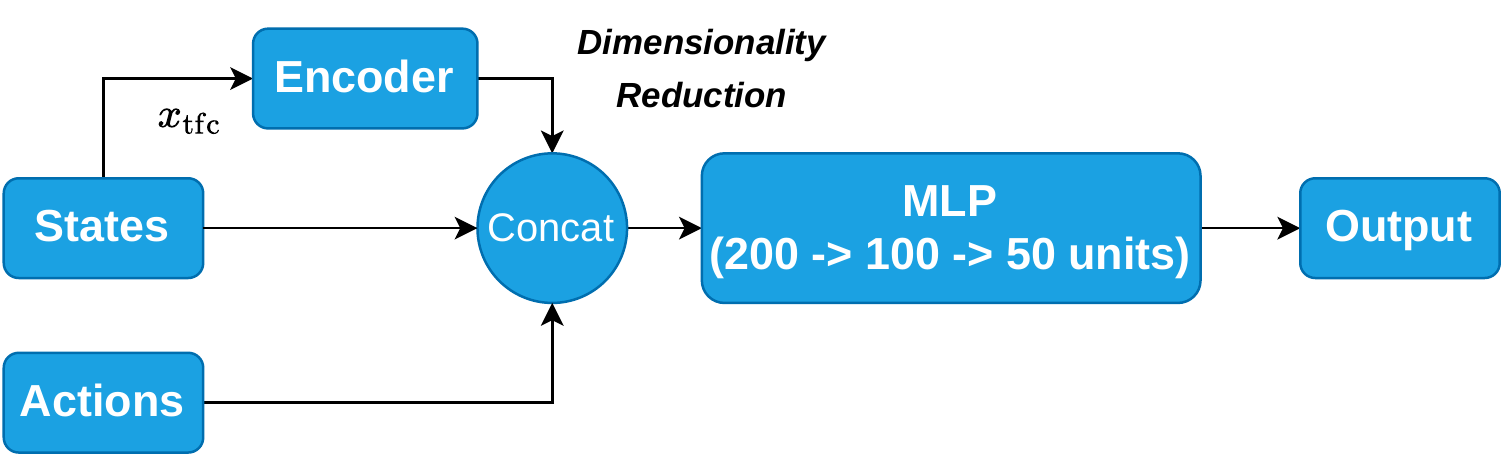}
    \caption{The Network Architecture of the Value and Q Functions.}
    \label{fig: value_graph}
\end{figure}

Fig. \ref{fig: value_graph} shows the architecture of the neural network associated with the Q-learning. Upon receiving the state vector, the sampled traffic light status $w_{\mathrm{tfc}}$ are fed to a pre-trained autoencoder with Multilayer Perceptron (MLP) of size $(81, 100, 5, 100, 81)$ for dimensionality reduction. 
The remaining states along with the actions are concatenated with the latent states from the encoder and subsequently fed into another MLP of size $(200, 100, 50)$ to output the Q function for critic and the perturbation for the actor. The critic and actor do not share parameters in this work. 

To accelerate the training and improve generalizability, the state of the vehicle is randomized for every 50 steps in simulation. When the domain randomization occurs, the battery $SoC$ and the vehicle velocity $v_{\mathrm{veh},t}$ are sampled from uniform distributions $\text{Uniform}(SoC^{\mathrm{min}}, SoC^{\mathrm{max}})$ and $\text{Uniform}(0, v_{\mathrm{lim},t})$, respectively. To guarantee feasibility during the trip, the domain randomization is disabled $200 m$ within signalized intersections or $1000 m$ within the final destination.  

\subsection{Safe Set Approximation}
In the eco-driving problem, two types of constraints can induce feasibility issues, namely, the battery terminal $SoC$ constraint and the constraint imposed by traffic rules at signalized intersections. 
For the first case, the goal set is considered not reached when the vehicle is near the destination and it cannot sufficiently charge the battery back to $SoC^\mathrm{T}$ in the remaining distance. 
For the second case, the trip is considered failed when the vehicle breaks traffic rules at signalized intersections, and infeasibility occurs when the vehicle speed is too high and there is not enough distance to brake to stop in front of a traffic light in red or stop sign.  
A conservative low-speed controller that only uses ICE is used to collect the initial data for the experience replay buffer.
During training, only samples from the trips that reach the goal set without violating any constraints are added to the experience replay buffer. 

In the eco-driving problem, the sampled traffic light $x_{\mathrm{tfc}}$ is binary, while the other variables are continuous. 
As the PDDP solver also discretizes the continuous state space, we consider the loss of accuracy with the same discretization is acceptable. 
The discretized states as in one-hot form in each dimension are fed into an LSTM network with 50 units sequentially as shown in Fig. \ref{fig: autoregressive_lstm}. The outputs from LSTM are then masked according to the number of categorical classes in each dimension. Finally, the softmax operator ensures the outputs to be a proper conditional probability distribution.
As an alternative to LSTM, Causal 1D Convolutional Neural Networks (Causal Conv1D) \cite{oord2016wavenet} was also implemented as the network for the autoregressive model. 
The key difference is that the states in all dimensions can be fed into Causal Conv1D in parallel, whereas each dimension needs to be fed into LSTM sequentially. For applications with long sequences, Causal Conv1D can be more efficient and accurate \cite{bai2018empirical}.  
For the specific problem, Causal Conv1D shows no noticeable advantage over LSTM in either accuracy or inference speed.
As a result, LSTM is chosen as it has fewer hyperparameters. 

When the receding horizon ($200 m$) in this study is longer than the critical braking distance \cite{zhu2021deep}, the vehicle will never violate any constraints imposed by signalized intersections. 
Nevertheless, using the safe set to constrain the terminal state is still essential for the following reason. 
At the last step of the receding horizon, the value function $V^\pi(\tilde{x}_{t+\Delta t_{H_s}})$ needs to be evaluated numerically for trajectory optimization. 
Since only the data from the safely executed trips are added into the buffer and there is no penalty mechanism for the constraint violation, the estimation of the critic network is valid only within the safe set and is subject to extrapolation error outside. 
Although the long receding horizon ensures the feasibility of the actual trajectory regardless of the use of a safe set, the training is subject to instability and the learned performance can significantly deteriorate without the constraint from the safe set.

\begin{figure}[!t]
    \centering
    \includegraphics[width=1\columnwidth]{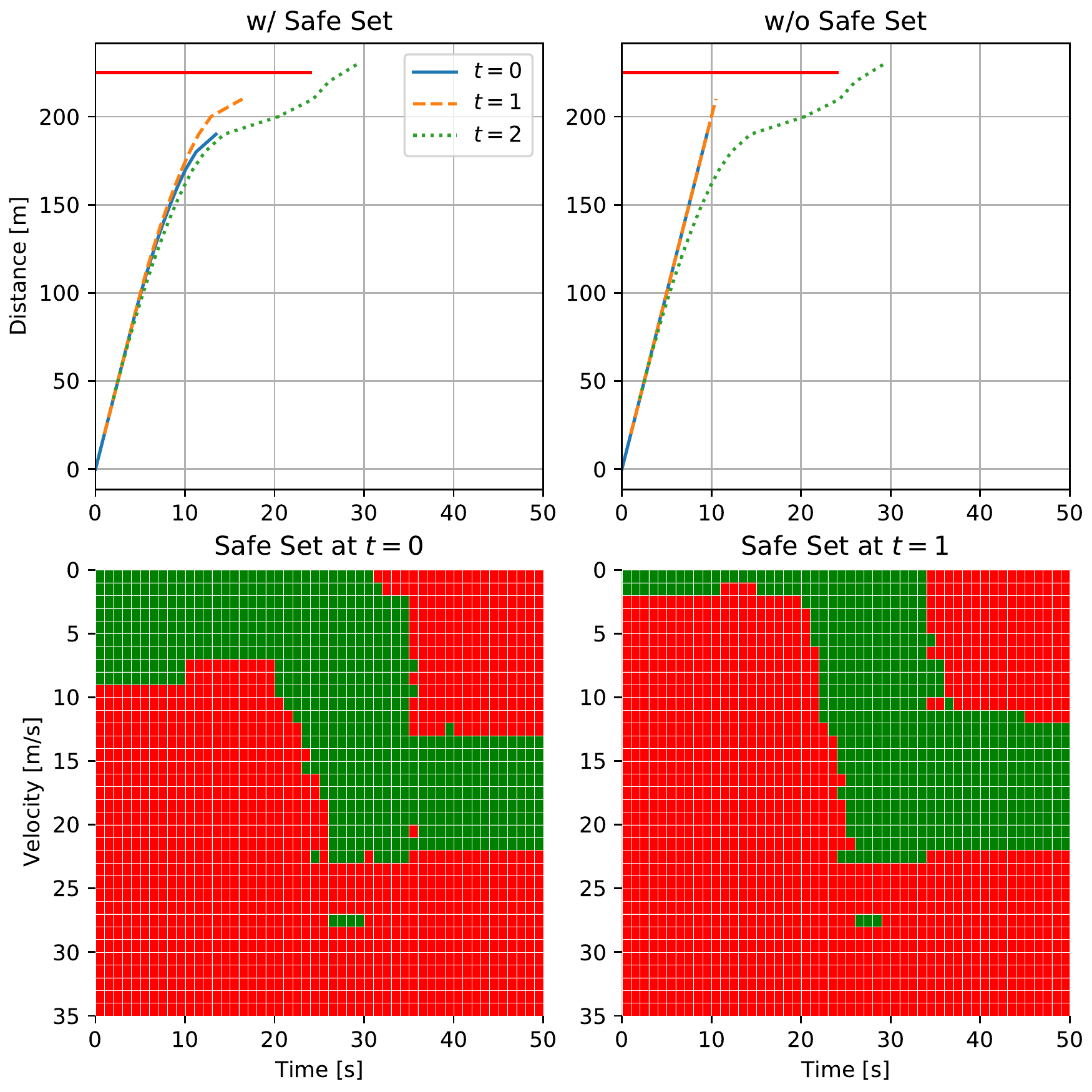}
    \caption{The Effect Of the Safe Set on Trajectory Optimization.}
    \label{fig: safe_set_effect}
\end{figure}

This effect is shown in Fig. \ref{fig: safe_set_effect}. 
Here, the two subplots on top show the optimized trajectories with and without the use of a safe set, respectively. 
The three curves in each plot are the trajectories from the optimizer at three consecutive seconds. 
During the first two seconds, the vehicle is more than $200m$ away from the traffic light, and thus the constraint from Eqn. \eqref{eq: traffic_constraint} is not considered in the trajectory optimization. 
The subplots on the bottom show the safety status of the terminal state in the dimension of the vehicle velocity and the time at which it reaches the end of the receding horizon before the signalized intersection appears in the receding horizon. 
Here, green means the state is considered safe, i.e. $p_\psi(x) \geq \delta$, and red otherwise. 
Although the actual trajectories, with or without a safe set, can slow down in time to avoid trespassing the red light thanks to the sufficiently long receding horizon, the terminal state without the safe set constraint has a speed of $20 m/s$ with $20 m$ left before a red light, which is unsafe.
Meanwhile, comparing the bottom two subplots, the terminal velocity constrained by the safe set progressively reduces as the vehicle approaches the intersection in the red phase.
In addition, given speed limit here is set to $v_\mathrm{lim}=22 m/s$, any state with a velocity higher than $22 m/s$ is considered unsafe.
It can be noticed that the red region on the top right corners is incorrectly considered unsafe (false positive). 
This is because, by optimality, the agent rarely crosses an intersection in the green phase with low speed, therefore, these false positive regions do not affect the performance.

As a summary for all the implementation details, the hyperparameters are listed in Tab. \ref{tab: hyperparameters}. 

\begin{table}[]
\caption{Hyperparameters of the Q Learning and Safe Set Estimation}
\label{tab: hyperparameters}
\centering
\begin{tabular}{l|l}
Parameter                      & Value \\ \hline
Weighting factor between fuel and time, $\lambda$ & 0.45 \\
Discount factor, $\gamma$                & 0.995      \\
Optimizer                      & Adam \\
Learning rate, $\alpha$                 & 1e-4      \\
Experience buffer size         & 2e5      \\
Batch size, $N$                     & 256     \\
Target network update rate, $\tau$     & 1e-3      \\
Exploration rate, $\epsilon$               & 0.2      \\
Perturbation range in physical unit, $\Phi$      & 30 Nm   \\
Sampled actions from the VAE decoder, $n$& 10 \\
Steps per domain randomization & 50      \\ 
LSTM size for the safe set & 50 \\
\end{tabular}
\end{table}

\section{Results}\label{sec: results}
\begin{figure}[!t]
    \centering
    \includegraphics[width=1\columnwidth]{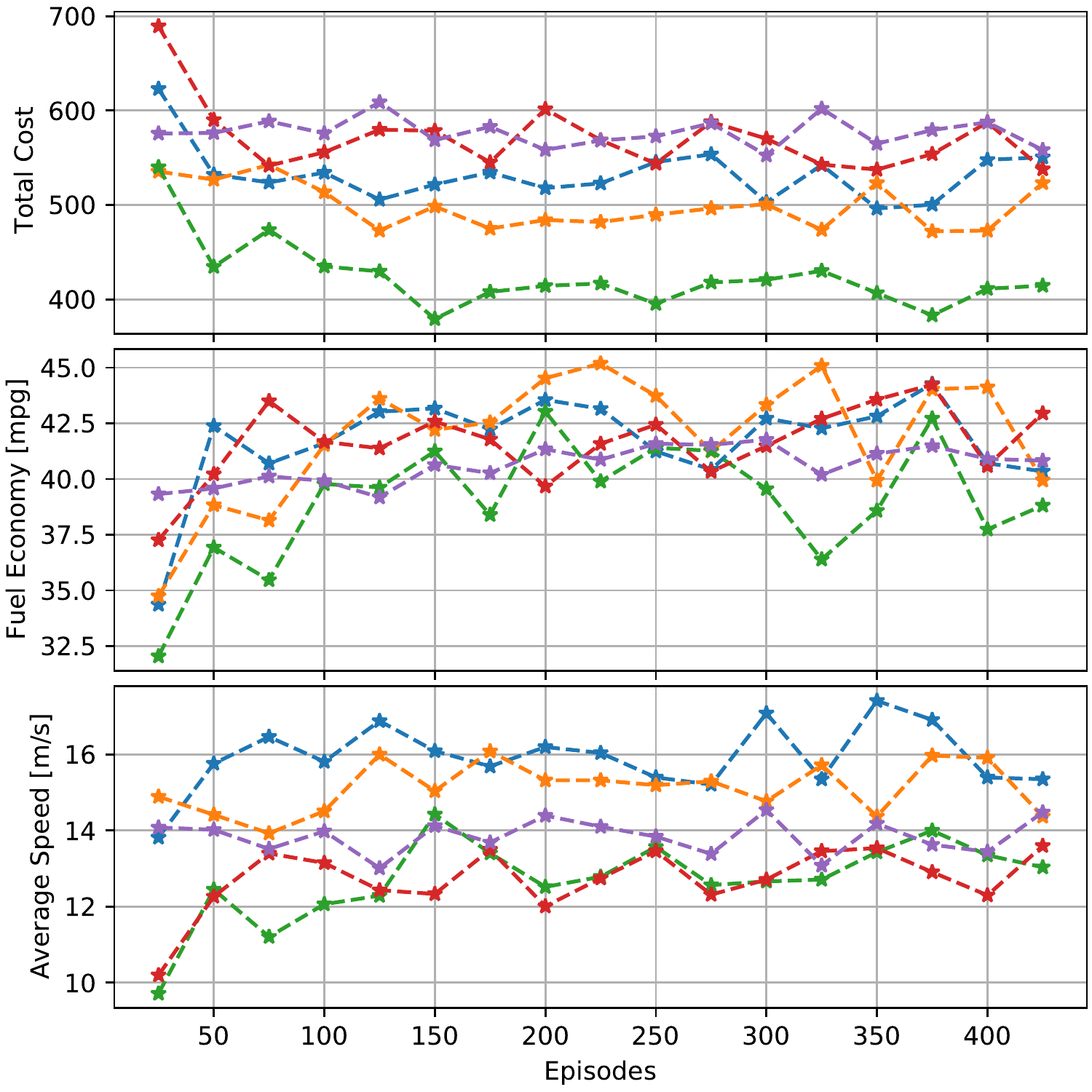}
    \caption{The Evolution of the Total Costs, Fuel Economy, and Average Speed of the 5 Evaluation Trips.}
    \label{fig: learning_curve}
\end{figure}

Both the PDDP optimizer and the neural network training require GPU.
To get the results to be shown, the training took 24 hours on a node with an NVIDIA Volta V100 GPU and $2.4 GHz$ Intel CPU from Ohio Supercomputer Center \cite{OhioSupercomputerCenter1987}.
As domain randomization is used during training, 5 trips out of 1000 randomly generated trips are repeatedly selected for every 25 training episodes to evaluate the performance of the controller and to quantify the progress of training. 
During the evaluation, domain randomization and epsilon greedy are both deactivated. Fig. \ref{fig: learning_curve} shows the evolution of the total costs, fuel economy and average speed of the 5 evaluation trips. 
Compared to the model-free on-policy method in \cite{zhu2021deep}, which takes 80,000 episodes to converge, the sample efficiency of the off-policy model-based method is significantly improved. 
In the meantime, with the constrained optimization formulation and the safe set, the training quickly learns to respect the constraints imposed by the terminal $SoC$ and signalized intersections. 
Furthermore, the fact that the agent does not need any extrinsic penalty from constraint violation and is still capable of learning to operate within the safe region significantly simplifies the design and tuning process as deployed in the previous reinforcement learning attempts on eco-driving \cite{zhu2021deep,pozzi2020ecological, li2019ecological}. 

Statistically, the performance of the agent trained with SMORL is compared against the four other strategies, a baseline strategy with Enhanced Driver Model (EDM) representing a human driver and heuristically calibrated energy management module \cite{gupta2019enhanced}, the hierarchical optimal controller using Approximate Dynamic Programming (ADP) in \cite{deshpande2021real}, the model-free DRL (MFDRL) agent proposed in \cite{zhu2021deep} and the wait-and-see (WS) solution.
The WS solution assumes the speed limits and the sequences of all traffic lights of the entire trip are known \textit{a priori}, and it is solved by PDDP solver as well.
Despite being non-causal and computational intractable for online implementation, the solution serves as the upper bound for the causal control strategies. 
The four strategies are evaluated on the 100 random trips as shown in Fig. \ref{fig: map_of_columbus}, and the fuel economy, average speed and variance of the battery $SoC$ are listed in Tab. \ref{tab: statistical_comparison}.

\begin{table}[]
    \centering
    \caption{Fuel Economy, Average Speed and SoC Variance for Baseline, Model-free DRL, SMORL and WS Solutions}
    \label{tab: statistical_comparison}
    \begin{tabular}{c|c|c|c|c|c}
    \multicolumn{1}{l|}{} & \multicolumn{1}{c|}{Baseline} & \multicolumn{1}{c|}{ADP} & \multicolumn{1}{c|}{MFDRL} & \multicolumn{1}{c|}{SMORL} & \multicolumn{1}{c}{WS} \\ \hline
    \begin{tabular}[c]{@{}c@{}}Fuel Economy\\ mpg\end{tabular} & 32.4 & 39.5 & 40.8 & 41.6 & 47.5\\ \hline
    \begin{tabular}[c]{@{}c@{}}Speed Mean\\ $m/s$\end{tabular} & 14.1 & 13.9 & 12.5 & 14.0 & 14.5\\ \hline
    \begin{tabular}[c]{@{}c@{}}$SoC$ Variance\\ $\%^2$\end{tabular}  & 12.1 & 21.6 & 18.2 & 52.6  & 22.6\\ 
    \end{tabular}
\end{table}

Here, compared to the baseline strategy, the SMORL agent consumes $21.8 \%$ less total fuels while maintaining a comparable average speed.
The benefit in fuel economy is achieved by avoiding unnecessary acceleration events and by taking advantage of a wider range of battery capacity as indicated by the higher $SoC$ variance. 
The SMORL agent shows dominant performance in average speed and fuel economy compared to the previously trained MFDRL strategy and the non-learning-based ADP strategy.
The performance improvement over MFDRL is primarily due to two aspects. 
First, the online trajectory optimization solved by PDDP guarantees the global optimality within the receding horizon, which is more accurate and reliable than the actions generated from the one-step stochastic policy from neural networks as in MFDRL. 
Second, the fact that there is no extrinsic penalty to assist constraint satisfaction ensures that the agent focuses on learning only the objective of the OCP formulation, i.e. weighted sum of the trip time and fuel consumption, instead of a carefully designed yet delicate surrogate learning objective.

\begin{figure}[]
    \centering
    \includegraphics[width=1\columnwidth]{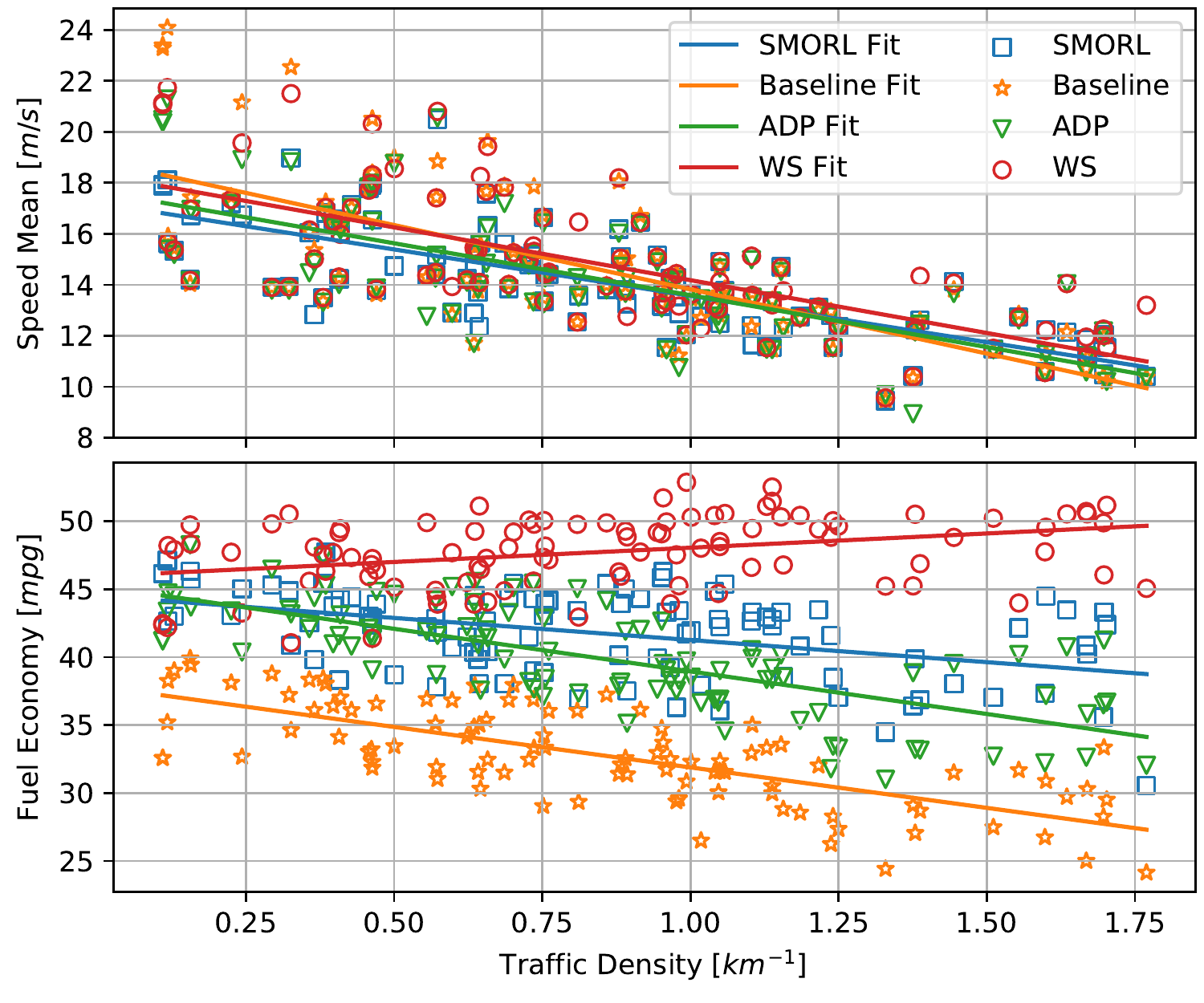}
    \caption{The Variation of the Average Speed and the Fuel Economy against Traffic Light Density for Baseline, SMORL and WS Solution}
    \label{fig: density plot}
\end{figure}

In Fig. \ref{fig: density plot}, the average vehicle speed and the fuel economy of each trip are plotted against the traffic light density. 
As the WS solution calculates the global optimal solution with the knowledge of the full trip, it is able to navigate among the traffic lights accordingly, as indicated by the surprisingly increasing fuel economy.  
This can be due to the fact that when there are more traffic lights, the vehicle is forced to operate with a lower speed and lower fuel consumption condition.
On the other hand, as the baseline driver has limited line-of-sight \cite{gupta2019enhanced} and the ADP, MFDRL and SMORL controllers have limited DSRC sensing range, the fuel economy decreases as the traffic light density increases. 
Nevertheless, as indicated by the slope of the fitted curve, the fuel economy of SMORL is less affected by the increase of the traffic light density compared to the baseline and ADP controller. 

\begin{figure}[]
    \centering
    \includegraphics[width=1\columnwidth]{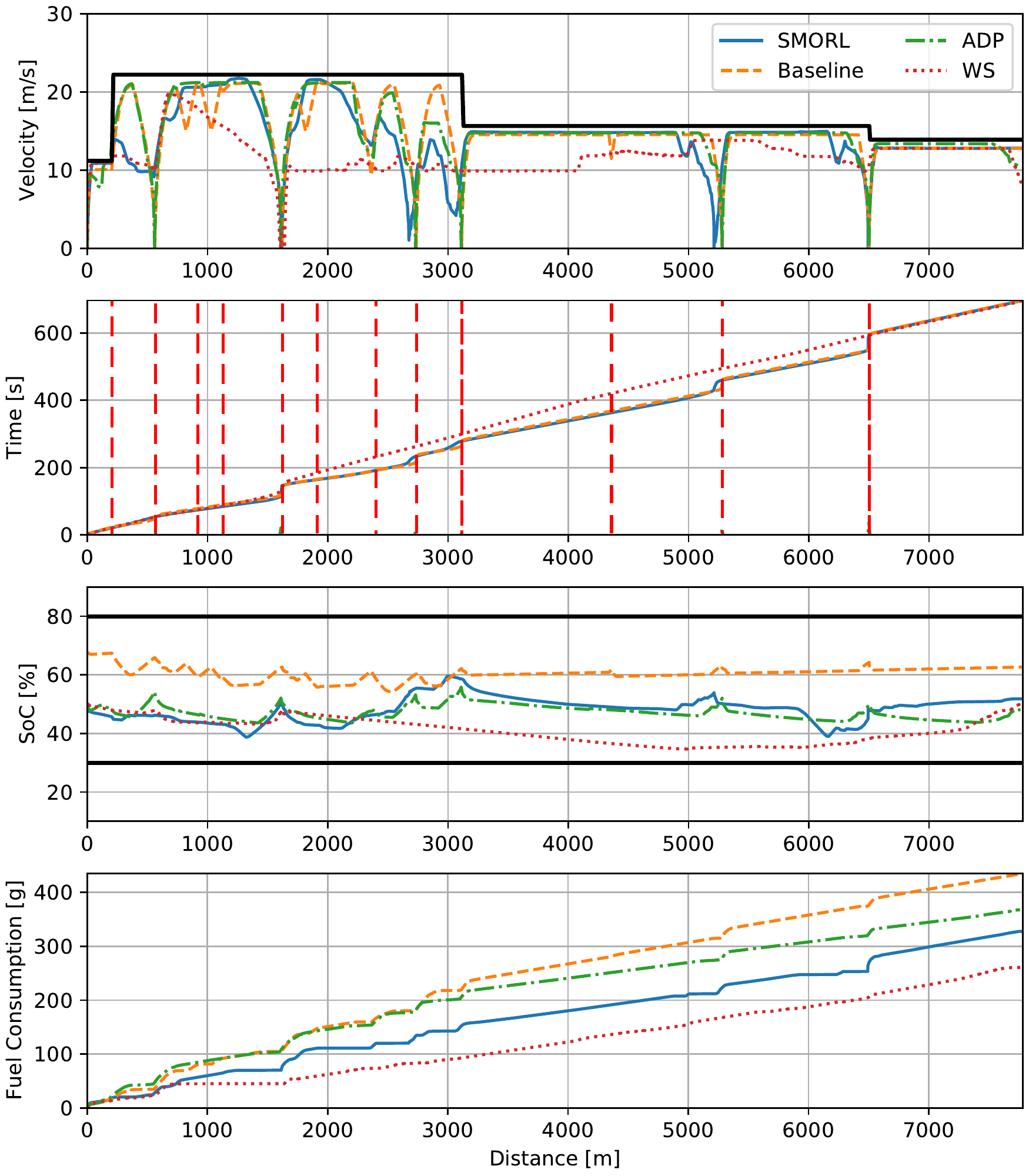}
    \caption{The Trajectory Comparison among Baseline, SMORL and WS.}
    \label{fig: comparison individual trip}
\end{figure}

Fig. \ref{fig: comparison individual trip} shows the comparison among the baseline, ADP, SMORL and the WS solution on a specific testing trip. 
For this specific trip, while the differences in trip time are within $3s$, SMORL consumes $24.7\%$ and $11.0\%$ less fuel compared to the baseline and the ADP strategies, respectively.
While SMORL demonstrates some merits similar to the WS solution, its inferiority to the WS solution is primarily due to the fact that only the SPaTs from the upcoming intersection are available to the controller. 
Additional comparisons among the four strategies can be found in Appendix \ref{appendix: additional comparison}.
The ablation study for the key components in the algorithm is shown in Appendix \ref{sec: ablation}. 

\section{Conclusion}\label{sec: conclusion}
In this paper, a safe-critical model-based off-policy reinforcement learning algorithm SMORL is proposed. The algorithm is applied to the eco-driving problem for CAHEV. 
Compared to the previous model-free attempts on eco-driving in the literature, the method does not require any extrinsic rewarding mechanism, and thus, greatly simplifies the design process and improves the final performance.
With the online constrained optimization formulation and the approximate safe set, the learned strategy is capable of satisfying the constraints in the prediction horizon and restricting the state within the approximate safe set, which is an approximation to the robust control invariant set. 
The performance of the strategy trained with SMORL is compared to a baseline strategy representing human drivers' behavior over 100 randomly generated trips in Columbus, OH, US. With a comparable average speed, the strategy from SMORL consumes approximately $22\%$ less fuel. 

While the demonstration of the algorithm is on the eco-driving problem, we believe it can be applied to many other real-world problems, in particular to those with well-studied system dynamics, such as robotics and autonomous driving. 
Future studies include the extending the SMORL algorithm to include the presence of leading vehicles, as well as the integration and verification of the algorithm in a demonstration vehicle.

\appendices
\section{Variational Autoencoder}\label{appendix: vae}
Let $X = \left\lbrace x_i \right\rbrace_{i=1}^N$ be some data set and $Z$ represent a set of low-dimensional latent variables, the objective is to maximize the marginal log-likelihood:
\begin{equation}
\begin{aligned}
    \log p(X) = \sum_{i=1}^N \log p(x_i) &= \sum_{i=1}^N \int_z \log p(x_i|z)p(z) dz\\
    &= \sum_{i=1}^N \mathbb{E}_{z\sim p(z)} \log p(x_i|z),
\end{aligned} \label{eq: marginal loglikelihood}
\end{equation}
As Eqn. \eqref{eq: marginal loglikelihood} is in general intractable, its variational lower bound is instead maximized:
\begin{align}
    \begin{aligned}
    \mathcal{L}(\omega_1, \omega_2, X) =  & -D_\mathrm{KL}\left(q_{\omega_1}(z|X)|| p(z)\right) \\
    & + \mathbb{E}_{q_{\omega_1}(z|X)} \left[\log p_{\omega_2}(X|z)\right] 
    \end{aligned}. \label{eq: vae update}
\end{align} 
Here, $D_{\mathrm{KL}}$ is the Kullback–Leibler (KL) divergence, and $p(z)$ is the prior distribution that is typically assumed to be a multivariate normal distribution. 
$q_{\omega_1}(z|X)$ is the posterior distribution parametrized by $\omega_1$.
To analytically evaluate the KL divergence, the posterior is typically constructed as $\mathcal{N}(z|\mu_{\omega_1}(X), \Sigma_{\omega_1}(X))$.
From a coding theory perspective, $q_{\omega_1}(z|X)$ and $p_{\omega_2}(X|z)$ can be considered as a probabilistic encoder and a probabilistic decoder, respectively. 

To compute the $\nabla_{\omega_1} L(\omega_1, \omega_2, X)$, policy gradient theorem \cite{williams1992simple} or Reparametrization trick \cite{williams1992simple, kingma2013auto} can be used. The latter is often used in VAE as it typically leads to a lower variance. 

In practice, the encoder $q_{\omega_1}(z|X)$ and the decoder $p_{\omega_2}(X|z)$ can be any function approximator. 
An implementation of VAE as the generative model to sample actions can be found in \url{https://github.com/sfujim/BCQ}. In this work, the latent space dimension is selected to be 5, and the encoder and the decoder are both MLPs with 2 layers of 300 hidden units.

\section{Autoregressive Model with LSTM}\label{appendix: autoregressive lstm}
\begin{figure}[!t]
    \centering
    \includegraphics[width=1\columnwidth]{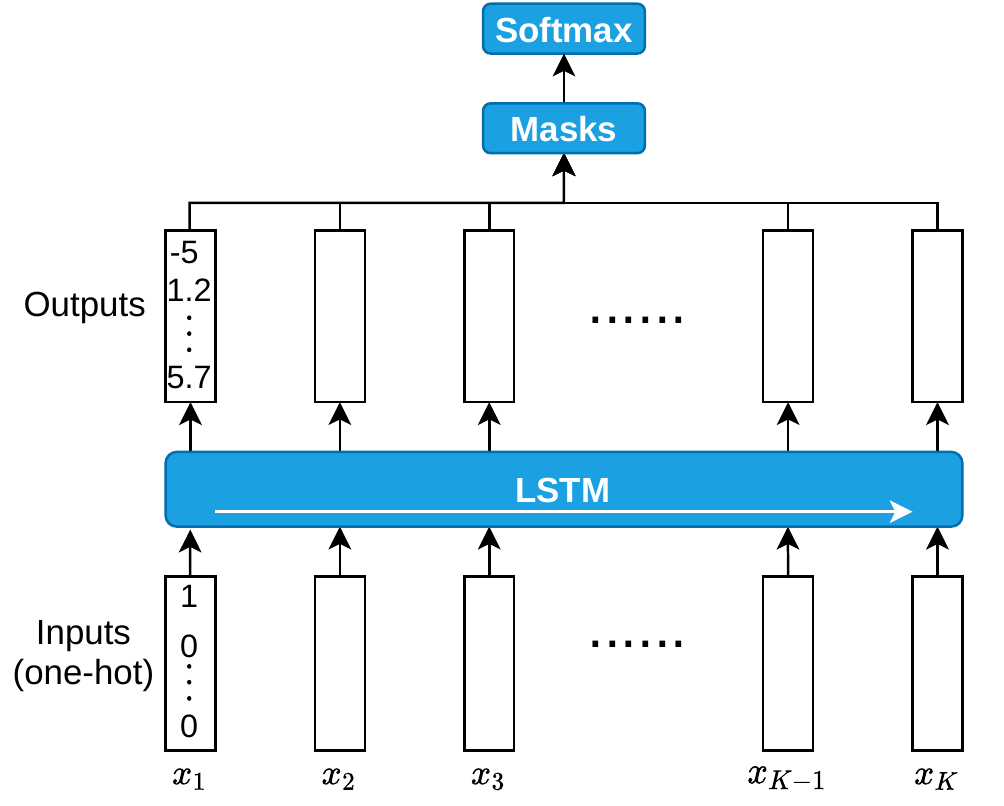}
    \caption{The Network Architecture of Recurrent Autoregressive Model.}
    \label{fig: autoregressive_lstm}
\end{figure}

For any probability distribution, the joint distribution can be factorized as a product of conditional probabilities as follow:
\begin{equation}
\begin{aligned}
    &p(x) = \prod_{i=1}^K p(x^{(i)}|x^{(1)}, ..., x^{(i-1)}), \\
    \rightarrow &\log p(x) = \sum_{i=1}^K \log p(x^{(i)}|x^{(1)}, ..., x^{(i-1)}),
\end{aligned}
\end{equation}
where $x^{(i)}$ is the $i^{th}$ dimension of the discrete input vector and $K$ is the dimension of the input vector.
As shown in Fig. \ref{fig: autoregressive_lstm}, the input vector in one-hot vector form is fed to the LSTM network in sequence. 
The $i^{th}$ output of the LSTM network after the softmax operation becomes a proper conditional probability $p(x^{(i)}|x^{(1)}, ..., x^{(i-1)})$. The model is trained by minimizing the KL divergence between the data distribution sampled from the experience replay buffer and the modeled distribution:
\begin{equation}
\begin{aligned}
    & \min_{\psi} \: D_{\mathrm{KL}}\left[p^*(x)||p_\psi(x)\right]\\
    =& \min_{\psi} \: \mathbb{E}_{x\sim p^*(x)}\left[-\log p_\psi(x) \right] + \text{constant} \label{eq: autoregressive update}
\end{aligned}
\end{equation}

In this work, the LSTM network has a single layer and 50 hidden size.

\section{Additional Comparison among Strategies} \label{appendix: additional comparison}
\begin{figure}[!t]
    \centering
    \includegraphics[width=1\columnwidth]{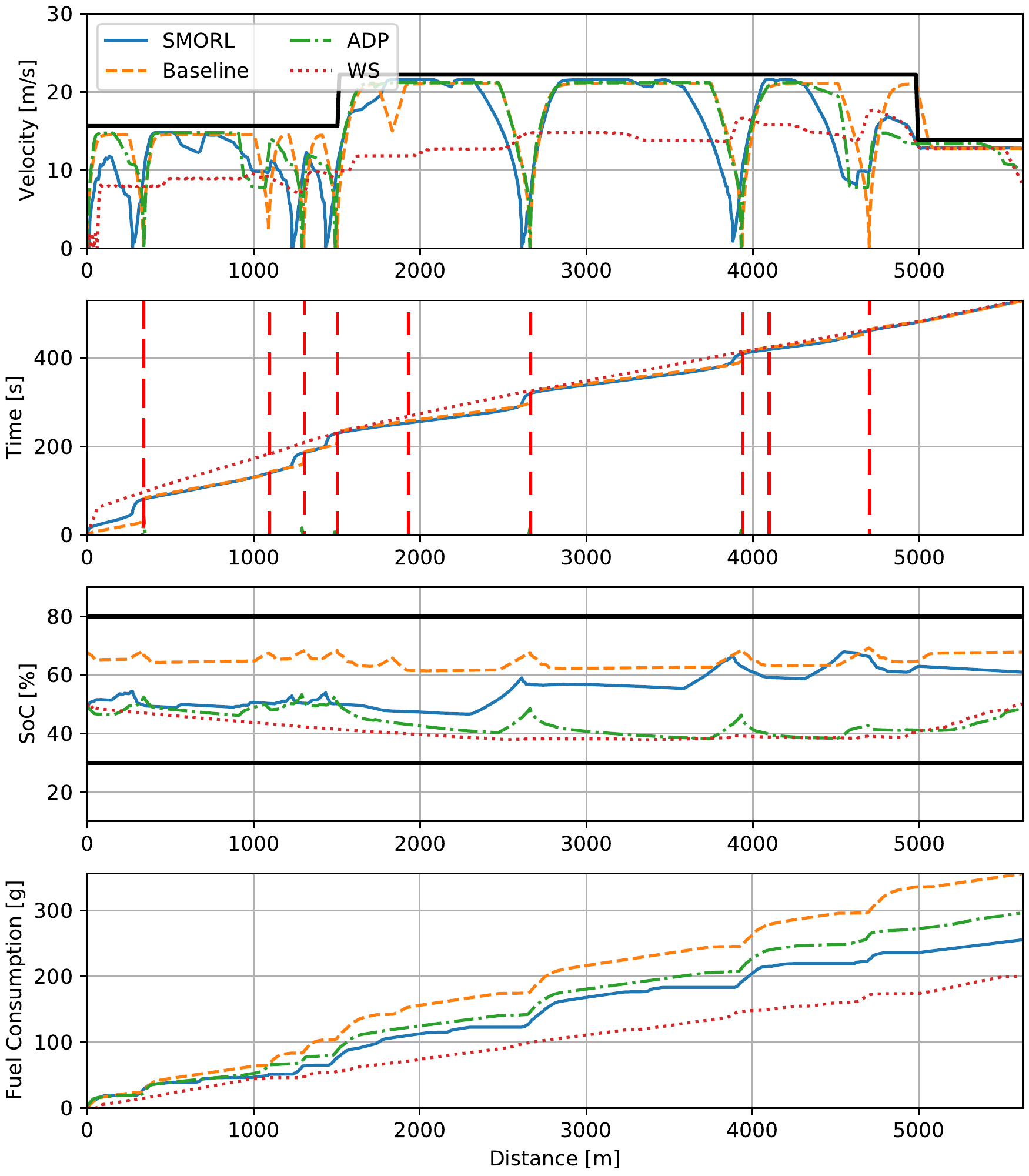}
    \caption{Comparison for High-density Low-speed Scenario.}
    \label{fig: comparison individual trip high density}
\end{figure}

\begin{figure}[!t]
    \centering
    \includegraphics[width=1\columnwidth]{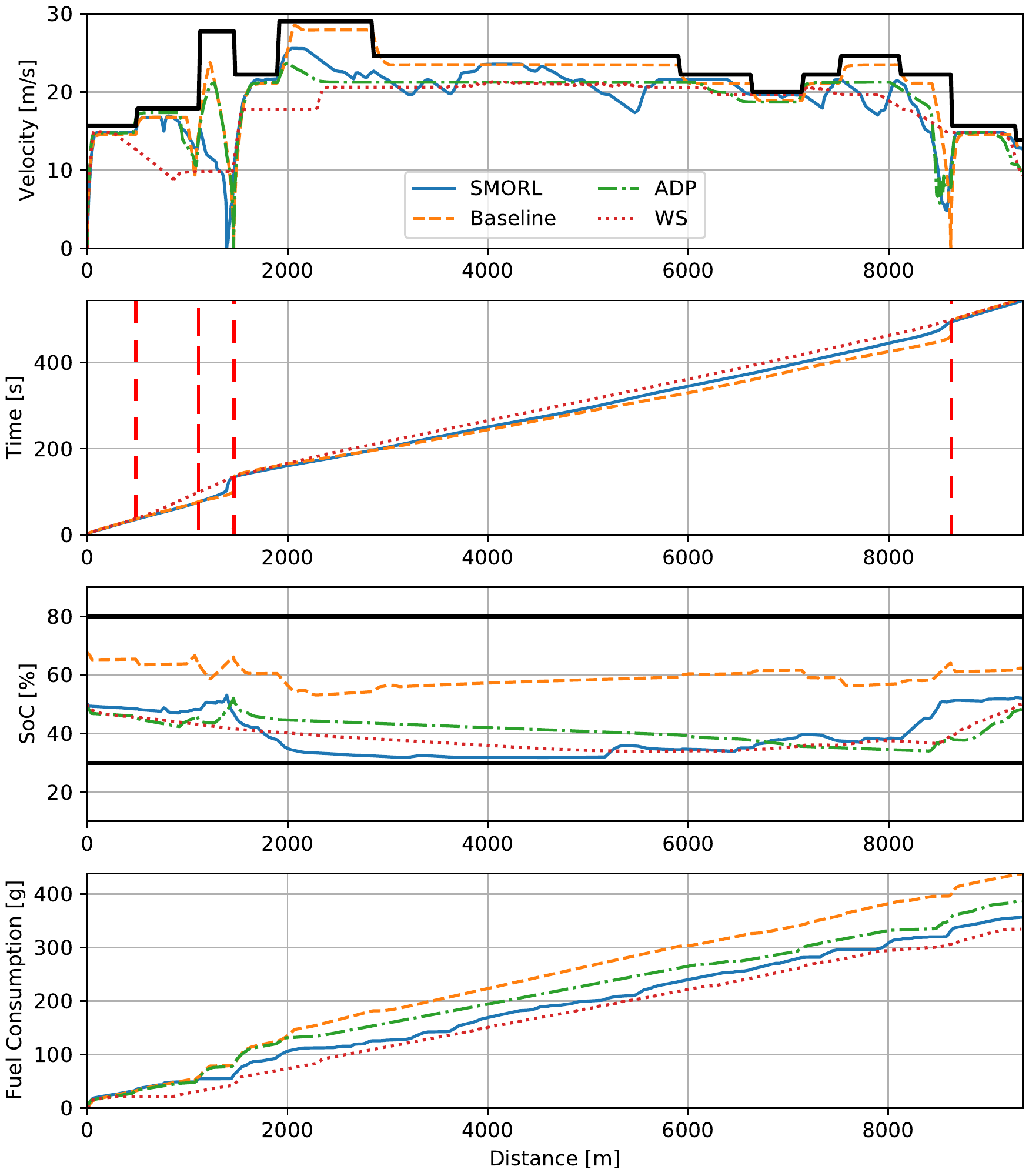}
    \caption{Comparison for Low-density High-speed Scenario.}
    \label{fig: comparison individual trip low density}
\end{figure}

Here, we show the comparison on two additional trips. 
The trip shown in Fig. \ref{fig: comparison individual trip high density} contains a large number of signalized intersections. 
As indicated by Fig. \ref{fig: density plot}, the gap between SMORL and the baseline and the gap between the wait-and-see solution and SMORL are both amplified by the high traffic density. 
The trip shown in Fig. \ref{fig: comparison individual trip low density} has a very low traffic density and the speed limits higher.
In such case, the difference between SMORL and the wait-and-see solution becomes less noticeable. Meanwhile, SMORL was still able to consume less fuel by using the capacity of the battery more efficiently.  

\section{Ablation} \label{sec: ablation}
In this part, we compare the full SMORL algorithm with the four intermediate algorithms. 
All the algorithms presented below use trajectory optimization solved via the PDDP solver. 
Tab. \ref{tab: ablation} shows the difference in configuration and compares the trained final performance over the 100 trips used for testing. 
Here, we see that the safe set and BCQ both have a positive impact on the trained performance. 
In fact, the combination of TD3 and trajectory optimization (Config. 2) does not provide any show any significant improvement over trajectory optimization only (Config. 1). 
In addition, without the use of the safe set, the controller will deplete the battery $SoC$ to $SoC^\mathrm{min}$ at the end of the trip as the terminal state constraint cannot be considered unless with the help of extrinsic penalty. 

\begin{table}[]
    \caption{Ablation Study for SMORL}
    \label{tab: ablation}
    \centering
    \begin{tabular}{c|c|c|c|c|c}
     &
      \begin{tabular}[c]{@{}c@{}}Safe\\ Set\end{tabular} &
      Q-learning &
      \begin{tabular}[c]{@{}c@{}}Fuel\\ Economy\\ $mpg$\end{tabular} &
      \begin{tabular}[c]{@{}c@{}}Average\\ Speed\\ $m/s$\end{tabular} &
      \begin{tabular}[c]{@{}c@{}}Normalized\\ Cost\end{tabular} \\ \hline
    1 &                               & None & 43.1 & 11.2 & 100    \\ \hline
    2 &                               & TD3  & 39.1 & 13.9 & 88.0\\ \hline
    3 & \CheckmarkBold & TD3  & 39.9 & 13.3 & 90.5 \\ \hline
    4 &                               & BCQ  & 38.5     & 14.3     & 87.2     \\ \hline
    5 & \CheckmarkBold & BCQ  & 41.6 & 14.0 & 86.5
    \end{tabular}
\end{table}

\section*{Acknowledgment}
The authors acknowledge the support from the United States Department of Energy, Advanced Research Projects Agency-Energy (ARPA-E) NEXTCAR project (Award Number DE-AR0000794) and Ohio Supercomputer Center.

\ifCLASSOPTIONcaptionsoff
  \newpage
\fi

\bibliographystyle{IEEEtran}
\bibliography{references}

\end{document}